\documentclass[11pt,letterpaper]{article}

\usepackage{agenticlearning-pdflatex}
\usepackage{xspace}
\usepackage{etoolbox}
\usepackage{algorithm}
\usepackage[noend]{algpseudocode}
\usepackage{makecell}
\usepackage{pifont}

\newcommand{\ours}{sft-JEPA\xspace}
\newcommand{\adajepa}{AdaJEPA\xspace}
\newcommand{\jepattt}{JEPA-TTT\xspace}

\newcommand{\Cerr}{\mathcal{C}}
\newcommand{\Aerr}{\mathcal{A}}
\newcommand{\ebar}{\bar{\bm{e}}}
\newcommand{\vz}{\bm{z}}
\newcommand{\va}{\bm{a}}
\newcommand{\vb}{\bm{b}}
\newcommand{\ve}{\bm{e}}
\newcommand{\vg}{\bm{g}}
\newcommand{\vd}{\bm{d}}
\newcommand{\tz}{\theta_0}
\newcommand{\norm}[1]{\left\lVert #1 \right\rVert}

\newcommand{\std}[1]{\raisebox{-0.45ex}{\fontsize{5}{6}\selectfont$\pm$#1}}

\newcommand{\gr}[1]{\textcolor{black!75}{#1}}

\makeatletter
\DeclareRobustCommand\onedot{\futurelet\@let@token\@onedot}
\def\@onedot{\ifx\@let@token.\else.\null\fi\xspace}
\renewcommand{\eg}{\emph{e.g}\onedot}

\renewcommand{\etc}{\emph{etc}\onedot}
\newcommand{\vs}{\emph{vs}\onedot}
\newcommand{\wrt}{w.r.t\onedot}
\makeatother

\crefname{section}{Sec.}{Secs.}
\Crefname{section}{Section}{Sections}
\crefname{appendix}{Sec.}{Secs.}
\Crefname{appendix}{Section}{Sections}
\crefname{figure}{Fig.}{Figs.}
\Crefname{figure}{Figure}{Figures}
\crefname{table}{Tab.}{Tabs.}
\Crefname{table}{Table}{Tables}
\crefname{equation}{Eq.}{Eqs.}
\Crefname{equation}{Equation}{Equations}
\crefname{algorithm}{Algorithm}{Algorithms}
\Crefname{algorithm}{Algorithm}{Algorithms}

\hypersetup{citecolor=teal}

\makeatletter
\newsavebox{\wrap@box}
\newlength{\wrap@width}
\newcommand{\wrapprep}[3]{%
  \par\penalty\z@\setlength{\wrap@width}{#2}%
  \begin{lrbox}{\wrap@box}\begin{minipage}[t]{\wrap@width}\captionsetup{type=#1,hypcap=false}#3\end{minipage}\end{lrbox}%
  \ifdim\pagegoal=\maxdimen\else
    \ifdim\dimexpr\pagegoal-\pagetotal\relax<\dimexpr\ht\wrap@box+\dp\wrap@box+3\baselineskip\relax\newpage\fi
  \fi}
\newcommand{\wrapplace}{%
  \ifdim\pagegoal=\maxdimen\def\wrap@raise{0pt}\else\def\wrap@raise{-\intextsep}\fi
  \edef\wrap@lines{\the\numexpr(\dimexpr\ht\wrap@box+\dp\wrap@box+6pt+.5\baselineskip\relax)/\dimexpr\baselineskip\relax\relax}%
  \begin{wrapfigure}[\wrap@lines]{r}{\wrap@width}\vspace{\wrap@raise}\usebox{\wrap@box}\end{wrapfigure}}
\newcommand{\wrapclear}{\par\ifnum\c@WF@wrappedlines>\@ne\vskip\c@WF@wrappedlines\baselineskip\fi\WFclear}
\makeatother

\title{What Test-Time Training Corrects in Latent World Models? \\ A Slow-Fast Approach}
\shorttitle{sft-JEPA: Slow-Fast Test-Time Training for Latent World Models}

\author{%\small  \normalsize
  { Shuaicheng Niu$^{1}$, Guohao Chen$^{1}$, Kevin Z. Shen$^{2}$, Yunbei Zhang$^{3}$, \\ Peilin Zhao$^{4}$, Yinan Zhang$^{1}$, Cyril Leung$^{2}$ }\\[3pt]
  {\footnotesize $^{1}$Nanyang Technological University $^{2}$University of British Columbia $^{3}$Tulane University $^{4}$Shanghai Jiao Tong University} \\[1pt]
  {\normalsize\texttt{\{shuaicheng.niu\}@ntu.edu.sg} } \\ [1pt]
  {\href{https://sft-JEPA.github.io}{\texttt{sft-JEPA.github.io}}}
}
\shortauthor{Niu et al.}
\date{}

\makeatletter\providecommand\@LN[2]{}\makeatother

\begin{document}
\maketitle
\makeatletter\global\@topnum\z@\makeatother

\begin{abstract}
\vspace{-10pt}

Latent world models plan toward a goal by predicting the outcomes of candidate actions, but under distribution shift, prediction errors can lead the planner astray. Test-time training using executed actions helps, yet what it corrects and whether its updates should persist across episodes remain unclear. We address both by analyzing the world model's prediction error. On PushObj, an object-pushing task, we probe multiple actions from the same state and decompose the error into an offset shared across actions and an action-dependent component. The offset effectively shifts the planner's goal, and AdaJEPA mainly tracks it: the common shift alone recovers 91\% of its planning gain. Part of the offset is episode-specific, so carrying its correction into the next episode does not help and may even hurt. The remaining offset recurs across episodes and action-dependent error  can be or need to be learned and retained from many episodes. \ours therefore uses two learners on different timescales: a fast learner that tracks the offset and resets each episode, and a slow LoRA learner that accumulates recurring knowledge across episodes. Across PushObj, PointMaze, zero-shot planning on RoboCasa (with DINO-WM), and OGBench (with LeWAM), \ours leads all benchmark averages, improving planning success over \adajepa by up to 22 points.

\end{abstract}

\section{Introduction}
\label{sec:intro}

\begin{figure*}[t]
  \centering
  \includegraphics[width=\textwidth]{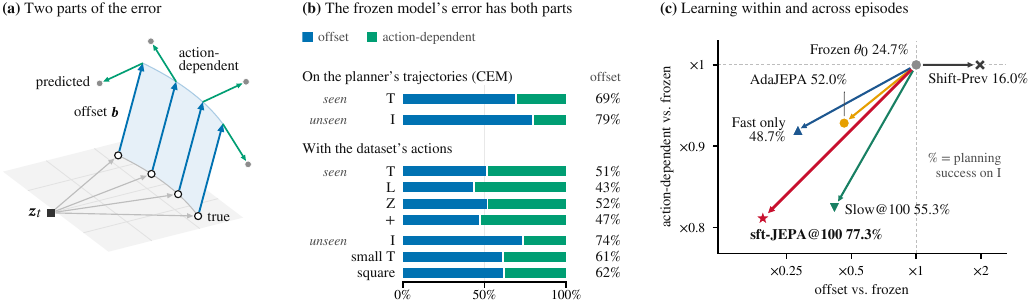}
  \caption{\textbf{Two kinds of test-time prediction error.}
  (a)~Illustration with four action chunks from the current latent $\vz_t$. Each prediction (dot) differs from the true outcome (ring) by an offset $\vb$ shared across actions (blue) and an action-dependent part (green).
  (b)~Composition of the frozen model $\tz$'s error energy on PushObj.
  (c)~Offset and action-dependent energies of each method relative to $\tz$ on PushObj (shape I), along with planning success rates.
  \adajepa and our fast learner alone (Fast only) mainly reduce the offset.  Reusing the previous episode's shift (Shift-Prev) enlarges the offset. Compared with \adajepa, Slow@100 (trained on 100 earlier episodes) further reduces the action-dependent error but only slightly reduces the offset. Our \ours{}@100 further reduces both parts and achieves a higher success rate.
 }
  \label{fig:teaser}
\end{figure*}

World models allow an agent to rehearse the future before acting~\citep{ha2018world,hafner2019planet,hafner2025dreamerv3,hansen2024tdmpc2}.
Joint-embedding predictive architectures (JEPAs)~\citep{lecun2022path,assran2023ijepa,bardes2024vjepa} further bring this idea into latent space, learning to predict representations rather than raw pixels.
Conditioned on actions, such latent predictors enable a planner to search for actions whose predicted latents lie closest to the representation of a goal image~\citep{zhou2025dinowm,sobal2025pldm,assran2025vjepa2,wang2026temporal}.
However, when the test environment differs from the training conditions~\citep{zhang2026liberomax}, \eg, in object shape, physical dynamics, or scene layout, even small prediction errors can distort the estimated distances to the goal and thus mislead action selection.

A natural solution is to keep learning during deployment, since every executed action reveals the true next observation as free signals. \adajepa~\citep{wang2026adajepa} implements this idea within model predictive control (MPC)~\citep{garcia1989mpc}: at each replanning step, it updates the model on recent transition windows and resets to the pretrained weights each episode, improving planning a lot. In contrast, \jepattt~\citep{zhang2026jepattt} retains what it learns across episodes through persistent learning. These choices motivate us to ask two questions:
Which changes in model predictions explain the planning gains of test-time training? Which corrections remain useful across episodes?

To answer both, we start from an action-fan probe that compares the predicted and true next latents of action chunks executed from the same simulator state (\cref{fig:teaser}a).
We split the prediction error into an \emph{offset}, shared across all actions, and an \emph{action-dependent} component. Then, in our PushObj (an object-pushing task) analysis, as in \cref{fig:teaser}b, the frozen model exhibits both error components, even for a shape seen in training. Based on this anatomy (\cref{sec:analysis}), we have:

$\bullet$ \emph{\adajepa mainly corrects the offset to help planning.} \textbf{(1)}~Since the planner scores candidates by their distance to the goal, the offset actually shifts a goal. Thus, correcting the offset can reduce the planner's optimistic prediction errors and improve planning success;
\textbf{(2)}~\adajepa mainly tracks this offset (\cref{fig:teaser}c):
its updates shift the predictions for all actions by nearly the same vector, which changes slowly within an episode. Removing this shift eliminates its planning gain, while adding it alone to the initial frozen model recovers 91\% of the gain.

$\bullet$ \emph{Across episodes, only corrections of recurring errors remain useful.} \textbf{(3)}~Part of the offset is \emph{episode-specific}: a correction fitted to one episode may not help the next and can even hurt.
Reusing a shift from the previous episode reduces planning success by 26 points compared to fitting it online, while a fast learner that is never reset loses 10--12 points;
\textbf{(4)}~The remaining offset recurs across similar situations and can be learned from multiple episodes. The \emph{action-dependent} error, barely reduced by within-episode adaptation, also requires cross-episode learning with more action-conditioned transitions.

Test-time adaptation thus has two jobs on different timescales: quickly tracking and then forgetting episode-specific offsets, while slowly learning and retaining what recurs across episodes. Inspired by these findings, we build our \ours method as follows. A \emph{fast} learner, consisting of the predictor's last block, tracks the offset after every chunk and resets at the start of each episode. A \emph{slow} learner, consisting of LoRA adapters on the whole predictor, learns what recurs from past episodes through a small update after each episode and retains this knowledge throughout deployment.

\paragraph{Main novelty and contributions.}
  \textbf{(1)} We introduce an action-fan probe that separates the prediction error in latent world models into an offset shared across actions and an action-dependent part. Together with controlled planning tests, it enables planning gains to be attributed to specific error corrections, providing a way to identify what test-time training actually learns.
  \textbf{(2)} We find that within-episode adaptation mainly corrects the prediction offset, with little effect on action-dependent error. However, part of this offset can be episode-specific, so retaining its correction may fail to help or even harm subsequent episodes, while the remaining offset recurs across similar situations and action-dependent errors can be or need to be learned across episodes.
  \textbf{(3)} Inspired by these, we propose a slow-fast test-time training method (\ours), which achieves the best average success on PushObj, PointMaze, RoboCasa (with DINO-WM) and OGBench (with LeWAM), outperforming \adajepa by 4--22 points.

\section{Preliminaries}
\label{sec:prelim}

\paragraph{Latent world model.}
A latent world model predicts the outcomes of actions in a learned latent space.
An encoder maps each frame $t$, an image possibly paired with proprioception, to a latent $\vz_t$.
Raw actions are grouped into \emph{chunks} of $k$ steps, and $\va_t$ is the chunk that starts at frame $t$.
An action-conditioned transformer predictor $f_\theta$ reads $L$ past latents spaced one chunk apart, each paired with the chunk that follows it, and predicts the latent one chunk ahead,
\begin{equation}
  \hat{\vz}_{t}=f_\theta\big(\vz_{t-kL},\dots,\vz_{t-k};\ \va_{t-kL},\dots,\va_{t-k}\big),
  \label{eq:predictor}
\end{equation}
where $L$ is at most the predictor's context length.
We write $\tz$ for the pretrained predictor.
A training \emph{window} $(t,L)$ is one such prediction of frame $t$ from $L$ context frames; it is \emph{chunk-aligned} when $t$ is a multiple of $k$.

\paragraph{Planning.}
Given the latent $\vg$ of a goal image, model predictive control (MPC)~\citep{garcia1989mpc} rolls out the predictor over several action chunks and searches for actions whose predicted latents approach $\vg$, measured by a weighted squared distance $\norm{\bm{v}}_W^2$ (\cref{sec:impl_plan}).
The planner for searching action chunks can be either the sampling-based cross-entropy method (CEM)~\citep{rubinstein2004cross}, or gradient descent (GD).
The agent then executes the first chunk, observes the outcome, and replans.

\paragraph{Test-time adaptation in the loop.}
Every executed chunk reveals the true next latents, providing a free supervision signal independent of how the actions were chosen.
\adajepa~\citep{wang2026adajepa} uses this signal after each executed chunk.
With a fresh Adam optimizer, it makes a small update to the predictor's last block, its final LayerNorm, and the encoder's projection head, using prediction errors from chunk-aligned windows over the last five chunks. The pretrained weights are restored at the start of each episode.
In contrast, \jepattt~\citep{zhang2026jepattt} continually fine-tunes the entire predictor and action encoder using an ever-growing replay buffer of past transition windows, retaining both the learned weights and historical data across episodes.

\section{Anatomy of Test-Time Prediction Error}
\label{sec:analysis}

Which changes in model predictions explain the planning gains, and which corrections remain useful across episodes?
To answer these, we study PushObj with the CEM planner, using a shape seen in training (T) and an unseen one (I), detailed in \cref{sec:supp_probe}.
For the first, we analyze the offset and its correction (\cref{sec:offset,sec:adajepa}).
For the second, we test which corrections transfer to later episodes (\cref{sec:specific,sec:dynamics}).

\begin{figure*}[t]
  \centering
  \includegraphics[width=\textwidth]{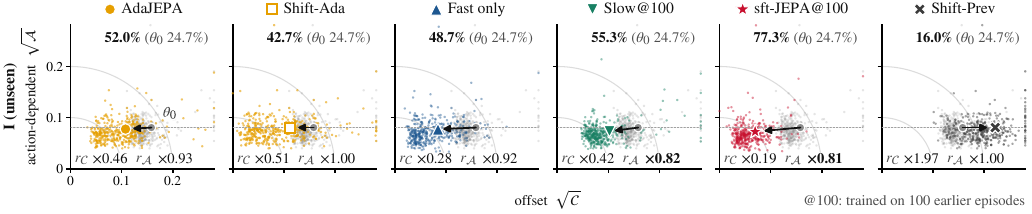}
  \caption{\textbf{What each method fixes.}
  Each dot is a probe state visited on the unseen PushObj shape I.
  Left/Down means a smaller offset/action-dependent error.
  Light gray dots show the frozen model $\tz$, the others show the method, and the arrow joins their means.
  Numbers give planning success and both error energies relative to $\tz$ ($r_{\Cerr}$ and $r_{\Aerr}$).
  Shift-Ada/Prev shift all predictions by one vector, fitted in the current or previous episode.
  In-episode adaptation moves the states mainly left. Moving them down needs training on many episodes.}
  \label{fig:plane}
\end{figure*}

\subsection{An action-fan probe}
\label{sec:probe}

A planner compares the predicted outcomes of many actions from the same state, so we measure the error the same way.
Along 150 episodes of the frozen model per shape, we stop at every replan, reset the simulator to that state, and execute a fan $G$ of nine action chunks: the dataset's mean action and four random pairs $\pm\va$ around it (\cref{sec:probe_fan}).
For each chunk, the error $\ve(\va)=\hat{\vz}(\va)-\vz^+(\va)$ compares the prediction with the true next latent (\cref{fig:teaser}a).

The \emph{offset} $\ebar=\frac{1}{|G|}\sum_{\va\in G}\ve(\va)$ is the average error over $G$: the part of the error that all actions share ($\vb$ in \cref{fig:teaser}a).
The rest of each error, $\ve(\va)-\ebar$, is its \emph{action-dependent} part.
Their energies (squared sizes) are the offset energy $\Cerr$ and the action-dependent energy $\Aerr$:
\begin{equation}
  \Cerr=\norm{\ebar}_W^2-\frac{\Aerr}{|G|},\qquad
  \Aerr=\frac{1}{|G|-1}\sum_{\va\in G}\norm{\ve(\va)-\ebar}_W^2.
  \label{eq:split}
\end{equation}
Here, the term $\Aerr/|G|$ corrects for the small fan: the average over nine actions differs from the average over all actions, which inflates $\norm{\ebar}_W^2$ by about $\Aerr/|G|$; with it, $\Cerr+\Aerr$ equals the mean error energy over $G$.
Adding the same vector to every prediction changes the offset but not the action-dependent part: each probe state is a point in the plane $(\sqrt{\Cerr},\sqrt{\Aerr})$, and a shift moves it only sideways (\cref{fig:plane}).

We replay every adaptive method along the same trajectories.
The probed models include \adajepa, closed-form shifts, and the two learners of \ours (\cref{sec:method}), each alone.
Fast only, like \adajepa, adapts the predictor's last block within an episode and is reset after it, but aims to track the offset more closely.
After every chunk, it takes a few Adam steps on dense windows, which use every frame of the newest chunk as a target.
Slow@100 is a LoRA trained on 100 earlier episodes and then frozen; it shows what a learner can carry over to episodes it has not seen.

\begin{table}[t]
\centering
\caption{\textbf{Controlled planning tests} on PushObj (CEM).
Success rate (\%) on the shapes T and I. Change: points against the model in the last column, T and I pooled.
-P: predictor only. Its shift: the part of \mbox{\adajepa-P's} change shared by all actions. \mbox{Shift-Online}: the frozen model's mean error on current windows, refitted after every chunk and subtracted from all predictions. \mbox{Shift-Prev}: the previous episode's final \mbox{Shift-Online} vector. @100: trained on its own stream's first 100 episodes, then frozen. \mbox{\ours{}@100} adds a fast learner reset every episode.}
\label{tab:controlled}
\footnotesize
\setlength{\tabcolsep}{2.3pt}
\begin{tabular*}{\linewidth}{@{\extracolsep{\fill}}lcccc@{\hspace{1.6em}}lcccc@{}}
\toprule
Planning model & T (seen) & I (unseen) & Change & vs. & Planning model & T (seen) & I (unseen) & Change & vs. \\
\midrule
\multicolumn{5}{@{}l}{\emph{Reference}} & \multicolumn{5}{l@{}}{\emph{Constant vs.\ learned, within vs.\ across episodes}} \\
Frozen $\theta_0$ & 48.7 & 24.7 &  & & Shift-Online & 72.0 & 40.7 &  & \\
\adajepa & 78.0 & 52.0 & $+$28.3 & $\theta_0$ & Shift-Prev & 45.3 & 16.0 & $-$25.7 & Shift-Online \\
\adajepa-P & 82.7 & 47.3 & $+$28.3 & $\theta_0$ & Fast only & 84.0 & 48.7 & $+$10.0 & Shift-Online \\
\midrule
\multicolumn{5}{@{}l}{\emph{Is the gain a shift?}} & \multicolumn{5}{l@{}}{\emph{Adding cross-episode knowledge}} \\
\adajepa-P without its shift & 47.3 & 22.7 & $-$1.7 & $\theta_0$ & Slow@100 & 76.0 & 55.3 &  & \\
$\theta_0$ + \adajepa-P's shift & 76.7 & 48.0 & $+$25.7 & $\theta_0$ & \ours{}@100 & 86.7 & 77.3 & $+$15.7 & Fast only \\
\bottomrule
\end{tabular*}

\end{table}

\subsection{The offset moves the goal}
\label{sec:offset}

On PushObj, the offset accounts for 43--79\% of the frozen model’s error across shapes (\cref{fig:teaser}b).
Correcting an offset error shared by all actions may look useless, since it leaves the differences between candidate actions unchanged.
However, the planner scores each candidate by the distance of its prediction $\hat{\vz}(\va)=\vz^+(\va)+\vb+\vd(\va)$ to the goal $\vg$, and
\begin{equation}
  \norm{\hat{\vz}(\va)-\vg\vphantom{\vz^+}}=\norm{\vz^+(\va)+\vd(\va)-(\vg-\vb)}.
  \label{eq:goalshift}
\end{equation}
Thus, an offset $\vb$ moves the goal to $\vg-\vb$ for any action-dependent part $\vd(\va)$.
With the offset uncorrected, the planner therefore steers the true state toward $\vg-\vb$ rather than $\vg$: it stops where the model sees the goal as reached, while the true state is off by the offset.
The model is thus optimistic about its own plan, as reported in reinforcement learning for policies or plans optimized on a learned model~\citep{kurutach2018model,janner2019trust}.
Correcting the offset moves the goal back to $\vg$ and alleviates this optimistic error, which is why a correction shared by all actions helps planning (\cref{sec:adajepa}).

\subsection{AdaJEPA mainly tracks the offset}
\label{sec:adajepa}

As shown in \cref{fig:plane}, taking the unseen shape I on PushObj as an example, \adajepa's update mainly reduces the offset, it halves the offset energy ($\times0.46$) but lowers the action-dependent energy only slightly ($\times0.93$).

For \adajepa-P, a predictor-only variant that performs on par with \adajepa, we define the \emph{shift} as the difference between its predictions and those of the frozen model at the current state, averaged over a fixed set of pre-defined actions and recomputed after every update (\cref{sec:probe_planning}).
Subtracting this shift from all of its predictions eliminates the planning gain (1.7 points below the frozen model, as shown in \cref{tab:controlled}).
Conversely, adding only this shift to the frozen model's predictions recovers 91\% of the gain (25.7 of 28.3 points, pooled over T and I; 82\% on T).
These results suggest that \adajepa acts mainly as an offset tracker in this setting.

\subsection{The episode-specific offset: track, then forget}
\label{sec:specific}

\begin{figure*}[t]
  \centering
  \includegraphics[width=\textwidth]{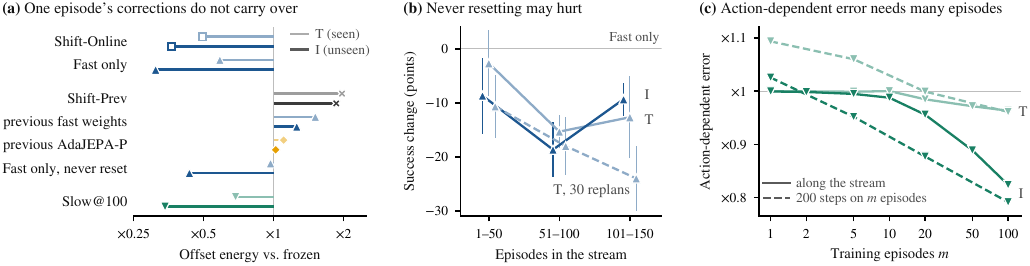}
  \caption{\textbf{What carries over across episodes.}
  (a)~Offset energy relative to the frozen $\theta_0$.
  Corrections fitted within the episode and Slow@100 lower it, whereas those carried over from the previous episode raise it or barely change it.
  A never-reset fast learner lowers it less than Fast only.
  (b)~Success of a never-reset fast learner relative to Fast only, in each 50-episode stage.
  With 30 replans (allowing more updates), the drop grows with every stage.
  (c)~Action-dependent error relative to the $\theta_0$ for a LoRA trained on $m$ episodes.
  T (light), I (dark).
  }
  \label{fig:timescales}
\end{figure*}

Within an episode the offset drifts slowly: the offsets of two states one chunk apart have a cosine of 0.87--0.89.
We compare two ways of tracking it, both updated after every chunk on the same windows:
1) Shift-Online, a closed-form shift that trains nothing, sets a single vector to the frozen model's mean error on these windows (an estimate of the current offset), and then subtracts it from all predictions;
2) Fast only instead takes a few Adam steps on the predictor's last block.
Both lower the offset energy of their own episode, to $\times0.31$--$0.59$ of the frozen model's (\cref{fig:timescales}a). Accordingly, both of them plan better than the frozen model, and Fast only outperforms Shift-Online by 10.0 points (\cref{tab:controlled}).

We then carry a correction unchanged into the next episode of the same shape: Shift-Prev keeps the previous episode's final Shift-Online vector, and we likewise keep the fast learner's final weights.
Both now raise the offset energy, to $\times1.86$--$1.97$ and $\times1.26$--$1.51$ of the frozen model's (\cref{fig:timescales}a), because a carried-over correction is nearly orthogonal to the one the new episode needs (cosine 0.05--0.10).
In planning, Shift-Prev falls 25.7 points below Shift-Online and even below the frozen model on both shapes (\cref{tab:controlled}), and a fast learner that is never reset loses 10.2 (T) and 12.2 (I) points over 150-episode streams (\cref{fig:timescales}b).

Part of the offset is therefore episode-specific and not shared across episodes. Thus, a correction fitted to this offset may not benefit other episodes and could even hurt.
A tracker should forget its correction when the episode ends.

\subsection{What transfers: learn slowly, then keep}
\label{sec:dynamics}

Not all of the offset is episode-specific, as it depends on the situation, and similar situations can recur across episodes.
For two states from different episodes, the closer they are, the more alike their offsets, \eg, the cosine is 0.57--0.71 for close pairs and 0.06--0.12 for distant ones.
Moreover, a correction fitted to one episode fails in the next because the two rarely visit the same states, and a learner that has seen many episodes can instead predict the offset from the input.
Slow@100, a LoRA trained on 100 earlier episodes and then frozen, never sees the test episodes, yet it lowers their offset energy to $\times0.42$ on I (\cref{fig:plane}) and $\times0.70$ on T.

The action-dependent error needs many episodes for a different reason.
Each episode contains at most 20 executed chunks, with only one action observed at each visited state, providing limited coverage of how prediction errors vary with the action.
This may explain why \adajepa's update barely reduces it (\cref{sec:adajepa}).
With more episodes, the error decreases (\cref{fig:timescales}c): on the unseen shape I, a LoRA trained along the stream remains at $\times0.99$ after 10 episodes but drops to $\times0.82$ after 100. With the same 200 training steps, a LoRA trained on 100 episodes achieves 23\% lower error than one trained on a single episode, suggesting that data diversity, rather than the training steps, drives the improvement.

Both therefore call for a learner that learns slowly, from many episodes, and keeps what it has learned.
Yet such a learner does not replace the tracker: on top of the LoRA of an \ours stream ($\times0.44$ on I after 100 episodes), the fast learner still lowers the offset energy to $\times0.19$ (\cref{fig:plane}).

\section{\ours: Slow-Fast Test-Time Training}
\label{sec:method}

\begin{figure*}[t]
  \centering
  \includegraphics[width=\textwidth]{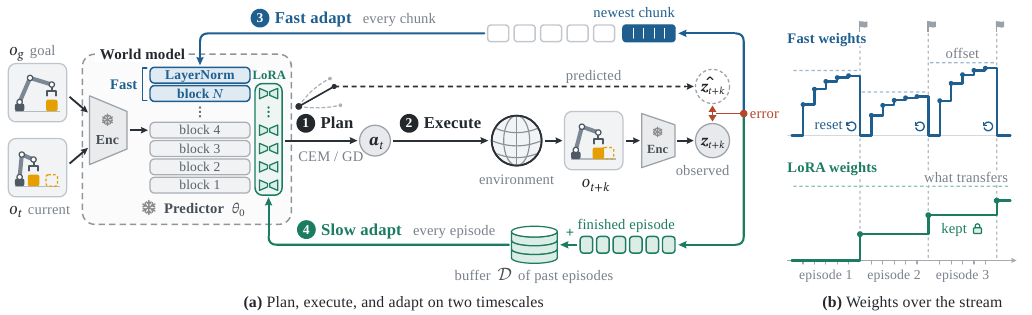}
  \caption{\textbf{Overview of \ours:} One world model, two learners on two timescales. The pretrained $\tz$ (gray) stay frozen. The \emph{fast} learner (blue) adapts the predictor's last block and final LayerNorm after every executed chunk and is reset when the episode ends.   The \emph{slow} learner (green), LoRA on every block, is updated after every episode on windows from past episodes and is kept. The curves are schematic.}
  \label{fig:method}
\end{figure*}

\cref{sec:analysis} points to two timescales: part of the offset is episode-specific (\cref{sec:specific}), whereas the rest of it recurs across episodes and the action-dependent error needs many episodes (\cref{sec:dynamics}).
This suggests that test-time training should
(i)~track the offset within the episode with a learned tracker;
(ii)~forget the tracker's correction when the episode ends; and
(iii)~learn what recurs, slowly and from many episodes, and keep what it has learned.
\ours follows these principles with two learners, a fast one for (i) and (ii) and a slow one for (iii).
How the error divides among these parts depends on the task, data and the model, so \ours always runs both learners rather than choosing between them.
They share one training signal but differ in what they update, which data they see and how long they live (as shown in \cref{fig:method} and \cref{alg:sft} in \cref{sec:impl_sft}).
The planner always uses $\tz+$LoRA$+$fast weights.

\paragraph{Training windows.}
Both learners change only the predictor and keep encoder frozen, so the latents are fixed targets.
They minimize the prediction error of windows $(t,L)$:
\begin{equation}
  \ell(t,L;\theta)=\norm{\hat{\vz}_t-\vz_t}^2 ,
  \label{eq:loss}
\end{equation}
where $\hat{\vz}_t$ is the prediction of \cref{eq:predictor} from the $L$ true context latents.
After chunk $n$ is executed, frames $0,\dots,kn$ are known.
\emph{Dense} windows use every frame of the newest chunk as a target ($t=kn-k+1,\dots,kn$), each with every context length.
\emph{Replay} windows add one chunk-aligned window from each of a few earlier chunks, the kind of window \adajepa uses.

\paragraph{Fast learner.}
The fast learner adapts the same parameters as \adajepa-P, the predictor's last block and final LayerNorm, to learn/track the offset within the episode.
After every executed chunk, a fresh Adam optimizer takes a few steps on the dense and replay windows, and the next replan uses the result.
These steps run on top of the current LoRA, so the fast learner tracks the offset that the slow learner has not removed.
At the end of every episode the fast learner is reset, so no episode inherits another's correction.

\paragraph{Slow learner.}
The slow learner keeps what transfers in LoRA adapters~\citep{hu2022lora} on all linear layers of the predictor; they start at zero and, with their Adam state, are kept for the whole deployment.
What transfers has to be learned from many episodes, so the slow learner draws on all past episodes.
When an episode ends, we reset the fast learner, add all dense windows of the episode to a buffer $\mathcal{D}$, and consolidate the slow learner with a few Adam steps on windows drawn uniformly from $\mathcal{D}$~\citep{rolnick2019experience}.
These steps run through $\tz+$LoRA; with only a few of them per episode, on windows from all past episodes, no single episode dominates the LoRA.

\section{Experiments}
\label{sec:exp}

\label{sec:setup}

\begin{table*}[t]
\centering
\caption{\textbf{Success rate (\%) on PushObj and PointMaze}.
Seen: T, L, Z, +; unseen: I, small T, square. Medium: the medium maze, averaged over default, low-mass and high-damping dynamics; layout: held-out layouts.
$^{\dagger}$Learns across episodes: its sd is over three orders of the same 150 episodes.
Gray rows show each of our two learners alone and are not ranked. See detailed results per task in \cref{sec:supp_bench} in Appendix.
Best in bold, second underlined.
}
\label{tab:main}
\footnotesize
\setlength{\tabcolsep}{2.6pt}
\begin{tabular*}{\textwidth}{@{\extracolsep{\fill}}l cccccc cccccc@{}}
\toprule
 & \multicolumn{3}{c}{PushObj (GD)} & \multicolumn{3}{c}{PushObj (CEM)} & \multicolumn{3}{c}{PointMaze (GD)} & \multicolumn{3}{c}{PointMaze (CEM)} \\
\cmidrule(lr){2-4} \cmidrule(lr){5-7} \cmidrule(lr){8-10} \cmidrule(l){11-13}
Method & Seen & Unseen & All & Seen & Unseen & All & Medium & Layout & All & Medium & Layout & All \\
\midrule
Frozen $\theta_0$ & 46.8\std{1.8} & 30.2\std{1.4} & 39.7\std{1.5} & 45.2\std{2.3} & 31.3\std{1.3} & 39.2\std{0.9} & 81.3\std{4.8} & 60.7\std{2.3} & 76.2\std{3.2} & 80.9\std{2.3} & 54.0\std{2.0} & 74.2\std{2.3} \\
\adajepa & \underline{81.0}\std{2.3} & 50.0\std{2.4} & \underline{67.7}\std{2.3} & \underline{74.2}\std{1.3} & 52.9\std{4.7} & \underline{65.0}\std{2.3} & \underline{83.6}\std{3.8} & 65.3\std{6.4} & 79.0\std{4.4} & 82.4\std{2.7} & 56.7\std{5.0} & 76.0\std{3.0} \\
\jepattt{}$^{\dagger}$ & 65.3\std{2.7} & 49.6\std{1.0} & 58.6\std{1.7} & 62.4\std{2.0} & 51.6\std{1.1} & 57.7\std{0.7} & 82.1\std{1.2} & 65.1\std{1.4} & 77.9\std{0.9} & \underline{83.3}\std{0.8} & 58.9\std{3.8} & 77.2\std{1.0} \\
\jepattt-EC$^{\dagger}$ & 71.1\std{3.4} & \underline{58.1}\std{1.6} & 65.5\std{2.6} & 66.7\std{1.5} & \underline{58.1}\std{2.6} & 63.0\std{1.9} & 81.6\std{0.8} & \textbf{80.7}\std{3.1} & \underline{81.3}\std{0.7} & \underline{83.3}\std{0.8} & \underline{74.2}\std{1.0} & \underline{81.1}\std{0.8} \\
\midrule
\textbf{\ours (ours)}$^{\dagger}$ & \textbf{82.9}\std{0.5} & \textbf{69.3}\std{1.0} & \textbf{77.1}\std{0.6} & \textbf{83.7}\std{0.6} & \textbf{70.7}\std{2.9} & \textbf{78.1}\std{0.9} & \textbf{84.9}\std{0.2} & \underline{77.3}\std{2.9} & \textbf{83.0}\std{0.6} & \textbf{85.5}\std{0.7} & \textbf{74.7}\std{2.3} & \textbf{82.8}\std{0.7} \\
\gr{\hspace{0.5em}$\bullet$ Fast only} & \gr{82.0\std{0.9}} & \gr{51.3\std{1.2}} & \gr{68.9\std{0.5}} & \gr{82.5\std{0.5}} & \gr{55.8\std{5.7}} & \gr{71.0\std{2.3}} & \gr{83.3\std{3.7}} & \gr{67.3\std{3.1}} & \gr{79.3\std{2.1}} & \gr{84.7\std{1.8}} & \gr{57.3\std{6.4}} & \gr{77.8\std{2.9}} \\
\gr{\hspace{0.5em}$\bullet$ Slow only$^{\dagger}$} & \gr{63.9\std{2.6}} & \gr{52.7\std{1.7}} & \gr{59.1\std{2.1}} & \gr{62.2\std{0.6}} & \gr{55.5\std{2.1}} & \gr{59.3\std{1.2}} & \gr{84.6\std{2.0}} & \gr{70.0\std{3.7}} & \gr{80.9\std{2.4}} & \gr{84.4\std{0.4}} & \gr{70.9\std{4.1}} & \gr{81.1\std{0.8}} \\
\bottomrule
\end{tabular*}
\end{table*}

\paragraph{Datasets and models.}
On PushObj and PointMaze, we use the world models and tasks released with \adajepa~\citep{wang2026adajepa}.
On PushObj, a variant of Push-T~\citep{chi2023diffusion}, the distribution shift is in the object's shape: the model is trained on four shapes (T, L, Z, +) and tested on these and on three unseen ones (I, small T, square).
On PointMaze~\citep{fu2020d4rl}, the shift is in the dynamics or the layout: a medium-maze model faces its default dynamics, a lower mass and a higher damping, and a model trained on 25 layouts faces five held-out ones.
In the simulated kitchens of RoboCasa~\citep{nasiriany2024robocasa}, the DINO-WM~\citep{zhou2025dinowm} from JEPA-WMs~\citep{terver2025jepawms}, trained on real-robot data~\citep{khazatsky2024droid}, plans zero-shot on the reach and place tasks.
LeWAM~\citep{fu2026lewam}, a world action model that predicts the next latent and the actions, plans with its released checkpoints on the Puzzle and Scene manipulation tasks of OGBench~\citep{park2025ogbench}.
To shift the dynamics, we rotate, scale or offset the robot arm's commands, or simulate a worn arm (see details in \cref{tab:other}).

\paragraph{Baselines.}
We compare with the frozen model, \adajepa and \jepattt~\citep{zhang2026jepattt}.
With the settings of its paper, \jepattt updates once for every 64 new windows, about once every 13 chunks.
We therefore add a variant, \jepattt-EC, which updates after every chunk and thus makes far more updates.
We also split \ours into its two learners, Fast only and Slow only.

\begin{table*}[t]
\centering
\caption{\textbf{Success rate (\%) on RoboCasa and OGBench (Puzzle and Scene)}.
DINO-WM~\citep{zhou2025dinowm} plans zero-shot in RoboCasa.
LeWAM~\citep{fu2026lewam} plans under No Shift and under four shifts of the arm.
R60 rotates the horizontal commands by 60$^\circ$.
WR$\pm$60 combines a worn arm (joint stiffness halved, joint friction added) with a rotation $[-60^\circ, 60^\circ]$, drawn in each episode.
O0.3 adds a constant offset of 0.3 to the horizontal commands (1.5\,cm per step).
G0.5+OE applies a gain of 0.5 to the commands and adds a horizontal offset drawn in each episode, of random direction and size up to 0.4.
Avg on LeWAM is calculated over the four shifts.
$^{\dagger}$Learns across episodes.
}
\label{tab:other}
\fontsize{7.5}{9}\selectfont
\setlength{\tabcolsep}{1.1pt}
\begin{tabular*}{\textwidth}{@{\extracolsep{\fill}}l ccc cccccc cccccc@{}}
\toprule
 & \multicolumn{3}{c}{RoboCasa (DINO-WM)} & \multicolumn{6}{c}{Puzzle (LeWAM)} & \multicolumn{6}{c}{Scene (LeWAM)} \\
\cmidrule(lr){2-4} \cmidrule(lr){5-10} \cmidrule(l){11-16}
Method & Reach & Place & Avg & No Shift & R60 & WR$\pm$60 & O0.3 & G0.5+OE & Avg & No Shift & R60 & WR$\pm$60 & O0.3 & G0.5+OE & Avg \\
\midrule
Frozen $\theta_0$ & 8.3\std{6.5} & 25.0\std{0.0} & 16.7\std{3.3} & 79.3\std{10.3} & 8.7\std{2.3} & 7.3\std{5.0} & 27.3\std{3.1} & 11.3\std{1.2} & 13.7\std{1.8} & \textbf{96.7}\std{1.2} & 22.0\std{7.2} & \underline{23.3}\std{2.3} & 55.3\std{10.3} & 28.0\std{5.3} & 32.2\std{2.3} \\
\adajepa & 13.5\std{6.5} & 25.0\std{6.3} & 19.3\std{6.3} & 78.0\std{8.7} & 30.0\std{6.0} & 10.7\std{5.0} & 52.7\std{9.9} & 28.7\std{1.2} & 30.5\std{5.1} & \textbf{96.7}\std{1.2} & 40.7\std{8.1} & 20.7\std{8.1} & 85.3\std{3.1} & \textbf{58.7}\std{8.3} & 51.4\std{4.5} \\
\jepattt{}$^{\dagger}$ & \underline{25.0}\std{3.1} & 31.3\std{3.1} & 28.2\std{2.7} & \underline{80.0}\std{6.9} & 15.3\std{6.1} & 12.7\std{5.0} & 58.7\std{8.1} & 15.3\std{5.8} & 25.5\std{4.3} & \textbf{96.7}\std{1.2} & 29.3\std{3.1} & 22.0\std{0.0} & 80.0\std{3.5} & 33.3\std{4.6} & 41.2\std{2.8} \\
\jepattt-EC$^{\dagger}$ & \underline{25.0}\std{19.5} & \underline{36.5}\std{4.8} & \underline{30.8}\std{9.9} & \textbf{80.7}\std{5.8} & \textbf{53.3}\std{8.1} & \underline{17.3}\std{4.2} & \underline{74.7}\std{6.4} & \underline{36.7}\std{3.1} & \underline{45.5}\std{4.8} & 94.0\std{2.0} & \underline{69.3}\std{8.1} & 22.0\std{3.5} & \underline{90.0}\std{2.0} & 51.3\std{10.1} & \underline{58.2}\std{2.8} \\
\midrule
\textbf{\ours (ours)}$^{\dagger}$ & \textbf{44.8}\std{1.8} & \textbf{38.5}\std{3.6} & \textbf{41.7}\std{2.4} & \underline{80.0}\std{6.9} & \underline{50.0}\std{10.6} & \textbf{30.7}\std{7.6} & \textbf{78.7}\std{5.8} & \textbf{45.3}\std{6.1} & \textbf{51.2}\std{7.1} & 96.0\std{0.0} & \textbf{76.0}\std{12.0} & \textbf{46.7}\std{1.2} & \textbf{93.3}\std{1.2} & \underline{58.0}\std{4.0} & \textbf{68.5}\std{2.6} \\
\gr{\hspace{0.5em}$\bullet$ Fast only} & \gr{19.8\std{4.8}} & \gr{22.9\std{9.5}} & \gr{21.4\std{2.4}} & \gr{78.7\std{9.5}} & \gr{40.0\std{3.5}} & \gr{29.3\std{5.0}} & \gr{62.7\std{8.3}} & \gr{42.7\std{3.1}} & \gr{43.7\std{4.5}} & \gr{94.7\std{3.1}} & \gr{59.3\std{2.3}} & \gr{44.0\std{7.2}} & \gr{92.0\std{2.0}} & \gr{62.7\std{8.1}} & \gr{64.5\std{3.1}} \\
\gr{\hspace{0.5em}$\bullet$ Slow only$^{\dagger}$} & \gr{29.2\std{13.0}} & \gr{27.1\std{7.9}} & \gr{28.2\std{2.7}} & \gr{79.3\std{6.4}} & \gr{40.7\std{6.4}} & \gr{14.7\std{7.0}} & \gr{73.3\std{5.8}} & \gr{31.3\std{5.0}} & \gr{40.0\std{4.4}} & \gr{97.3\std{2.3}} & \gr{59.3\std{2.3}} & \gr{18.7\std{2.3}} & \gr{90.7\std{1.2}} & \gr{48.7\std{8.1}} & \gr{54.4\std{1.9}} \\
\bottomrule
\end{tabular*}
\end{table*}

\paragraph{Implementation details.}
On PushObj and PointMaze, we use the planner settings (GD and CEM) released with \adajepa.
On RoboCasa, we follow the code and protocol of JEPA-WMs (CEM), but stop an episode at its first success.
LeWAM generates actions with its flow-matching policy, and we invert each method's adapted model to find the commands that realize them.
Our fast learner takes five Adam steps after every chunk, at a learning rate of $5\times10^{-4}$.
The slow learner, a rank-8 LoRA, takes two Adam steps after every episode, each on up to 256 windows, at a learning rate of $5\times10^{-4}$.
Methods that learn across episodes run on streams of 150 episodes on PushObj and PointMaze (three 50-episode stages), 32 on each RoboCasa task and 50 on LeWAM.
We report success rate as mean $\pm$ sd over three runs.
Full details are put in \cref{sec:supp_impl}.

\subsection{Main results}
\label{sec:results}

\paragraph{Results under Model Predictive Control (MPC).}
From \cref{tab:main}, we make four observations on PushObj and PointMaze.
1)~\ours achieves the highest success rate in 11 of 12 columns, outperforming \adajepa by 9.4/13.1 points on PushObj and 4.0/6.8 on PointMaze under GD/CEM planning, suggesting our effectiveness.
2)~On seen shapes, the pretrained model already predicts action dynamics fairly well, so within-episode offset tracking is effective. \adajepa and Fast only achieve 74.2--82.5\% across GD and CEM planners, outperforming persistent learners (62.2--71.1\%). Adding the slow learner further slightly improves \ours to 82.9/83.7\% under GD/CEM.
3)~On unseen shapes and held-out layouts, action dynamics are less accurate, making cross-episode learning more valuable. Thus, \jepattt-EC is the strongest except for \ours. However, learning across episodes alone is still insufficient: adding the fast learner gives \ours a clear advantage, \eg, 69.3/70.7\% \vs \jepattt-EC's 58.1/58.1\% under GD/CEM.
4)~\jepattt-EC outperforms \ours only on held-out layouts with GD. This may be because each layout recurs about 30 times per stream, providing repeated opportunities to learn its action dynamics. In this scenario, the more frequent updates of \jepattt-EC may thus give it an advantage over our slow learner, which updates only after each episode. Notably, our fast learner remains complementary to \jepattt-EC: adding it further improves success by 11.9/15.7 points on PushObj and 1.5/2.2 on PointMaze under GD/CEM, supporting the effectiveness of our proposed slow-fast learning mechanism (see \cref{sec:supp_tttfast} for more details).

\label{sec:other}

 We also examine zero-shot transfer on RoboCasa (\cref{tab:other}). From the results,
\adajepa provides limited gains, improving reach by only 5.2 points and showing no improvement on place, while cross-episode learning is more effective, as the domain gap and start--goal pairs recur along a stream.
\ours again achieves the highest success rates on both tasks, averaging 41.7\% compared with 30.8\% for \jepattt-EC.

\begin{figure}[t]
  \centering
  \includegraphics[width=\textwidth]{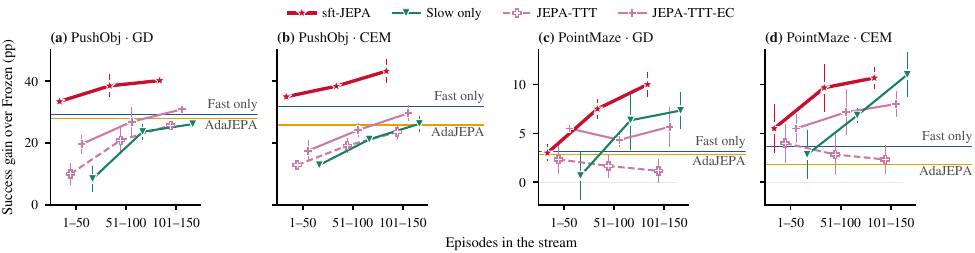}
  \caption{\textbf{Success gains along the stream.}
  Each marker shows the gain over the frozen model in a 50-episode stage, averaged across PushObj shapes (a, b) or PointMaze settings (c, d). Markers are shifted horizontally for clarity.
  Mean $\pm$ sd over 3 stream orders.
  }
  \label{fig:stream}
\end{figure}

\paragraph{Results on LeWAM (a world action model).}
We evaluate LeWAM under four shifts in the robot arm's dynamics. R60 and O0.3 remain fixed throughout a stream, while WR$\pm$60 and G0.5+OE also include a component that changes every episode.
From \cref{tab:other}, we have:
1)~\ours achieves the highest average success rate across the four shifts on both tasks, outperforming \adajepa by 20.7 and 17.1 points and \jepattt-EC by 5.7 and 10.3 points.
2)~Under fixed shifts, cross-episode learning is beneficial, as the same correction applies across episodes. Here, \jepattt-EC outperforms \adajepa by 4.7--28.6 points.
3)~Under WR$\pm$60, only \ours and Fast only show substantial improvements. \ours gains 23.4 points over the frozen model on both tasks, while no baseline gains more than 10.0 points. Since the rotation angle changes every episode, corrections retained from previous episodes may no longer apply, highlighting the importance of within-episode adaptation in this scenario.

\subsection{Ablations and more discussions}
\label{sec:ablation}
\label{sec:stream}

\paragraph{Effects of slow and fast learners.}
\ours outperforms both individual learners (as in \cref{tab:main,tab:other}) across all benchmark averages.
Fast only tends to perform better when pretrained action dynamics are relatively accurate or shifts vary across episodes, \eg, it achieves 82.0--82.5\% on seen PushObj shapes \vs 62.2--63.9\% for Slow only, and 29.3\%/44.0\% under WR$\pm$60 on Puzzle/Scene \vs 14.7\%/18.7\%.
Slow only can be more effective when shifts recur and action dynamics are less accurate, achieving 70.9\% on held-out layouts (CEM) \vs 57.3\% for Fast only.
Notably, combining the two can yield much larger gains than either alone, \eg, on unseen PushObj shapes, both achieve only 51.3--55.8\%, while \ours reaches 69.3\% (GD) and 70.7\% (CEM).
This highlights the benefit of combining our slow and fast learning.

\paragraph{Gains along the stream.}
\ours performs well from the start and improves further as the stream progresses (\cref{fig:stream}).
On PushObj, Slow only and both \jepattt variants gain only 8--20 points over the frozen model in the first 50 episodes, compared with 26--32 points for \adajepa and Fast only, suggesting the importance of sufficient within-episode offset correction.
\ours avoids this slow start with its fast learner, while its slow learner accumulates knowledge across episodes.
From the first to the last stage on PushObj, its lead over \adajepa widens from 5.4 to 12.2 points with GD and from 9.1 to 17.4 points with CEM.

\begin{figure}[t]
\begin{minipage}[t]{0.5\textwidth}
  \vspace{0pt}
  \makeatletter\def\@captype{table}\makeatother
  \centering
\caption{\textbf{Never resetting the fast learner} (Success rate (\%) on PushObj).
Reset: the fast learner restarts every episode from $\tz$.
Never: it keeps its change over the 150-episode streams.
Seen: T, L, Z, +; unseen: I, small T, square.
}
\label{tab:reset}
\scriptsize
\setlength{\tabcolsep}{2pt}
\begin{tabular*}{\linewidth}{@{\extracolsep{\fill}}ll ccc ccc@{}}
\toprule
 & & \multicolumn{3}{c}{GD} & \multicolumn{3}{c}{CEM} \\
\cmidrule(lr){3-5} \cmidrule(l){6-8}
Method & & Seen & Unseen & All & Seen & Unseen & All \\
\midrule
Fast only & Reset & 82.0\std{0.9} & 51.3\std{1.2} & 68.9\std{0.5} & 82.5\std{0.5} & 55.8\std{5.7} & 71.0\std{2.3} \\
 & Never & 69.3\std{1.3} & 43.8\std{1.7} & 58.4\std{1.2} & 69.5\std{1.7} & 46.5\std{1.2} & 59.7\std{0.5} \\
\midrule
\ours{} & Reset & 82.9\std{0.5} & 69.3\std{1.0} & 77.1\std{0.6} & 83.7\std{0.6} & 70.7\std{2.9} & 78.1\std{0.9} \\
 & Never & 70.1\std{2.2} & 48.8\std{2.2} & 61.0\std{2.2} & 70.2\std{1.6} & 52.1\std{0.6} & 62.4\std{1.1} \\
\bottomrule
\end{tabular*}

\end{minipage}\hfill
\begin{minipage}[t]{0.46\textwidth}
  \vspace{20pt}
  \centering
  \includegraphics[width=\linewidth]{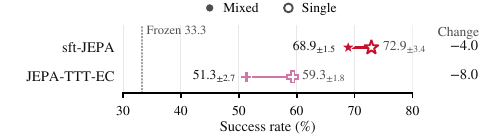}
  \caption{\textbf{Success rate (\%) on mixed PushObj shapes.}
  Mixed: a 150-episode stream alternating among T, I, square.
  Single: single-shape runs on the same episodes.
  Results under CEM planner.
  }
  \label{fig:mixed}
\end{minipage}
\end{figure}

\paragraph{Effects of resetting fast learner.}
The fast learner mainly tracks the offset, part of which is episode-specific, so \ours resets it at the start of each episode (\cref{sec:specific}).
\cref{fig:timescales}b showed that removing resets reduces success on shapes T and I.
\cref{tab:reset} extends this finding to all seven PushObj shapes, both planners, and \ours.
Without resets, Fast only loses 10.5 (GD) and 11.3 (CEM) points, while \ours loses 16.1 and 15.7 points, with declines on every shape.
PointMaze shows a slightly different pattern (\cref{sec:supp_reset}): removing resets even improves performance on held-out layouts, possibly because these layouts recur frequently within a stream, making carried-over fast weights useful across episodes.

\paragraph{Mixed shapes.}
We also evaluate streams that alternate among T, I, square across episodes (\cref{fig:mixed}).
Compared to single-shape streams on the same episodes, \ours drops by 4.0 points and \jepattt-EC by 8.0 points.
Nevertheless, \ours still outperforms the frozen model $\tz$ by 35.6 points and \jepattt-EC by 17.6 points.
Thus, \ours remains effective when shapes change across episodes.

\wrapclear
\wrapprep{figure}{0.5\textwidth}{%
  \centering
  \includegraphics[width=0.95\linewidth]{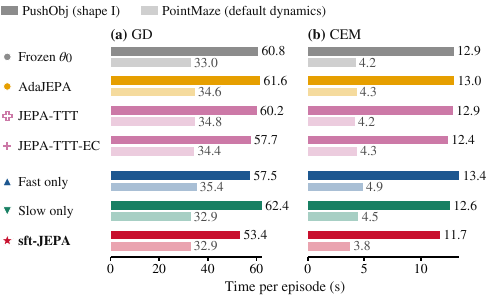}
  \caption{\textbf{Runtime per episode.}
  Mean wall-clock time per episode on PushObj (dark) and PointMaze (light).
  Each run covers 50 episodes on an RTX 3090 GPU.
  Episodes end at success, so methods that succeed earlier take less time.
  }
  \label{fig:cost}
}
\wrapplace
\paragraph{Runtime and memory costs.}
\ours is efficient in both runtime and memory overhead (\cref{fig:cost}).
The fast learner updates only the predictor's last block, while the slow learner takes two LoRA steps per episode.
Despite these updates, \ours takes no longer per episode than the frozen model, as the time saved by fewer replans (episodes end at success) exceeds the learning overhead.
Indeed, planning dominates runtime, \eg, each replan requires 3.4--3.8s (for GD) and 0.36--0.65s (for CEM) across tasks, while \ours spends only 0.06--0.07s for fast learning and 0.08--0.09s for slow learning between episodes.
Its peak allocated GPU memory exceeds that of the frozen model by 91--95 MiB, compared with 167--193 MiB for the \jepattt variants.
The slow learner's buffer of past episodes requires only 6.55--7.00 MiB (see \cref{sec:supp_buffer} in Appendix for more buffer design ablations).

\wrapclear

\section{Related Work}
\label{sec:related}

\paragraph{JEPA world models.}
Joint-embedding predictive architectures (JEPAs) learn representations by predicting in latent space rather than in pixel space~\citep{lecun2022path,assran2023ijepa,bardes2024vjepa}.
Conditioned on actions, they serve as world models that are trained on offline trajectories and reach goals with MPC~\citep{zhou2025dinowm,sobal2025pldm,assran2025vjepa2,wang2026temporal,terver2025jepawms} or action generation~\citep{fu2026lewam}.
These models are typically kept fixed at test time, even under distribution shift.
We study how to adapt them at test time from the transitions they observe.

\paragraph{Test-time adaptation}
updates pretrained models on unlabeled data through auxiliary self-supervision~\citep{sun2020ttt,liu2021tttpp,gandelsman2022tttmae,niu2026survey}, entropy minimization~\citep{liang2020shot,wang2021tent,zhang2022memo,niu2023sar,press2023rdumb}, teacher consistency~\citep{wang2022cotta,dobler2023rmt,yuan2023rotta,liu2024vida}, feature alignment~\citep{niu2024foa}, \etc.
In control, TARL~\citep{xu2025tarl} adapts policies by reducing action uncertainty, while PAD~\citep{hansen2021pad} uses an inverse-dynamics loss.
For world models, observed transitions provide a natural prediction target: \adajepa~\citep{wang2026adajepa} and \jepattt~\citep{zhang2026jepattt} both match predicted latents to encoded next observations.
They differ, however, in how updates are retained. \adajepa resets between episodes, whereas \jepattt learns persistently.
We analyze which components of prediction error these updates correct and which corrections should be retained or discarded across different episodes.

\paragraph{Slow-fast learning.}
Inspired by complementary learning systems~\citep{mcclelland1995why,kumaran2016what}, DualNet~\citep{pham2021dualnet} pairs a fast supervised learner with a slow self-supervised representation learner, while CLS-ER~\citep{arani2022clser} maintains plastic (fast) and stable (slow) models through moving averages at different timescales.
These methods accumulate knowledge across tasks, so what the fast learner acquires is kept.
In our \ours, the two timescales instead follow from our error analysis.
Corrections fitted within one episode may not transfer to the next, so the fast learner is deliberately reset after each episode, while the slow learner keeps what recurs across episodes.

\section{Conclusion}
\label{sec:conclusion}

In this paper, we investigate what adaptation corrects in latent world models and which corrections remain useful across episodes.
Using an action-fan probe, we decompose the prediction error into an offset shared across actions and an action-dependent component.
The offset shifts the planner's goal, and we find that \adajepa improves planning mainly by tracking this offset.
Moreover, part of the offset might be episode-specific, so carrying its correction into the next episode may not help and even hurt.
The remaining offset recurs across similar situations and can be learned across episodes, while action-dependent error, barely reduced by within-episode adaptation, also requires cross-episode learning with more action-conditioned transitions.
\ours therefore pairs a fast learner that tracks the offset and resets each episode with a slow learner that learns and retains recurring patterns from many episodes.
Across PushObj, PointMaze, RoboCasa, and OGBench, \ours improves planning success rate over \adajepa by 4--22 points.

{
    \small
    \bibliographystyle{ieeenat_fullname}
    \bibliography{main}
}

\clearpage
\appendix
\setcounter{table}{0}\renewcommand{\thetable}{\Alph{table}}\renewcommand{\theHtable}{S\arabic{table}}
\setcounter{figure}{0}\renewcommand{\thefigure}{\Alph{figure}}\renewcommand{\theHfigure}{S\arabic{figure}}
\setcounter{equation}{0}\renewcommand{\theequation}{\roman{equation}}\renewcommand{\theHequation}{S\arabic{equation}}

\noindent{\LARGE\sffamily\bfseries\color{LabDark} Appendix}\par\vspace{4pt}

This appendix has four parts.
\Cref{sec:supp_impl} describes the benchmarks, the world models, the baselines, \ours and the evaluation protocol in more detail.
\Cref{sec:supp_probe} details the error analysis of \cref{sec:analysis}.
\Cref{sec:supp_results} reports more results, per task and with other settings of the baselines.
\Cref{sec:supp_ablation} reports more ablations and discussions.

\section{More Implementation Details}
\label{sec:supp_impl}

The PushObj and PointMaze experiments run in the planning loop released with \adajepa~\citep{wang2026adajepa}, with its world models, environments, planners and evaluation code, on RTX 3090 GPUs with PyTorch 2.3.
Puzzle and Scene run on the same GPUs with PyTorch 2.11, and RoboCasa on A800 GPUs.

\subsection{Benchmarks}
\label{sec:impl_bench}

\paragraph{PushObj.}
The model is trained on four block shapes (T, L, Z, +) and tested on these and on three unseen ones (I, small T, square).
An episode starts from the first state of a 25-step segment of a test trajectory, and its goal is the observation reached by replaying the segment's actions.
Each shape has 1,000 such segments, and each seed draws 50 of them.
An episode succeeds when the 4-d vector of agent and block positions is within a Euclidean distance of 20 of the goal's (the workspace is $512\times512$) and the block angle is within $\pi/9$ of the goal's.
On PushObj and PointMaze, an episode ends at success or after 20 replans of five raw steps.

\paragraph{PointMaze.}
Maze cells have unit size.
The medium-maze model is tested with its default dynamics, a low mass and a high damping.
The last two scale the density of the point mass by 0.2 and the joint damping by 20.
Start and goal are drawn from the 26 open cells at a Euclidean distance of at least 3, with jittered positions and a random initial velocity.
The diverse-maze model is trained on 25 layouts and tested on five held-out ones.
The 50 episodes of a seed cycle through them, with a random start cell and a goal 3--5 cells away in breadth-first-search distance.
An episode succeeds when the point mass comes within 0.5 of the goal position.

\paragraph{RoboCasa.}
We use the reach and place tasks with the RoboCasa evaluation configurations of JEPA-WMs~\citep{terver2025jepawms}.
Start and goal come from task segments of demonstrations.
The goal is the dataset image at the end of its segment.
Reach succeeds when the hand is within 0.20\,m of the object and place when the object is within 0.15\,m of its target.
An episode lasts at most 60 raw steps and stops at its first success, checked after every five-step chunk, whereas JEPA-WMs runs all 60 steps and scores the final state.

\paragraph{Puzzle and Scene.}
Puzzle (3$\times$3) and Scene are goal-reaching tasks of OGBench~\citep{park2025ogbench}.
A robot arm presses buttons (Puzzle) or manipulates a cube, a drawer, a window and buttons (Scene) to match a goal image.
An episode starts at a random frame of a trajectory in the task's dataset, and its goal is the frame 25 raw steps later.
It succeeds when the task's objects match the goal and the end effector is within 4\,cm of its goal position.
It ends at success or after 50 raw steps (ten chunks).

We evaluate the model on its original task (No Shift) and the task under another four shifts (\cref{tab:other}).
They alter the commands or the arm itself.
The planner and the learners do not know about them.
By horizontal commands we mean the two horizontal components of the commanded end-effector translation.
R60 rotates the horizontal commands by 60$^\circ$.
WR$\pm$60 combines a worn arm (joint stiffness of all six joints halved, joint friction of 40\,N\,m, 7.5 at the wrist) with a rotation of the horizontal commands by $[-60^\circ, 60^\circ]$, drawn in each episode.
O0.3 adds a constant offset of 0.3 in normalized action units (1.5\,cm per step) to the horizontal commands.
G0.5+OE applies a gain of 0.5 to the translation commands and adds a horizontal offset of random direction and a size drawn from $U[0,0.4]$ (up to 2\,cm per step) in each episode.
The shifted commands are clipped to the action range.

\subsection{World models and planners}
\label{sec:impl_plan}
\label{sec:impl_wm}

\paragraph{\adajepa's models.}
We use the three \adajepa checkpoints (PushObj, medium maze, diverse maze), unchanged.
Each frame becomes one 404-d token of 384 visual, 10 proprioceptive and 10 action dimensions.
Losses use the 394 visual and proprioceptive dimensions.
The predictor is a six-block transformer (19.9M parameters, dropout 0.1, final LayerNorm) with frame-causal attention over up to three frames.
One model step spans one chunk of five raw steps ($k=5$ and $L\le3$ in \cref{eq:predictor}).
Planning rollouts are autoregressive.

\begin{table}[!htb]
\centering
\caption{\textbf{Planner settings on PushObj and PointMaze}, released with \adajepa and shared by all methods.
$\alpha$ is the weight of the proprioceptive term in the planning cost.
``All frames'' weights the current and the five predicted frames by $2^i$, normalized to sum to one.}
\label{tab:impl_planner}
\footnotesize
\setlength{\tabcolsep}{3pt}
\begin{tabular}{@{}>{\raggedright\arraybackslash}p{0.27\linewidth}>{\raggedright\arraybackslash}p{0.69\linewidth}@{}}
\toprule
Horizon, execution & 5 chunks (25 raw steps); execute the first chunk, then replan \\
Replanning & at most 20 replans per episode \\
Warm start & the 4 unexecuted chunks of the previous plan plus a new one \\
\midrule
GD & Adam on normalized actions, lr 0.1, 100 iterations, cosine decay to 0; new chunk at the zero action; no action noise \\
CEM~\citep{rubinstein2004cross} & 200 samples including the current mean, 30 elites, 10 iterations; $\sigma=1$ at every replanning step; mean warm-started, new chunk at 0 \\
\midrule
Cost, PushObj & $\alpha=1$; last frame only for the first 5 replanning steps, all frames afterward \\
Cost, medium maze & $\alpha=0$; all frames \\
Cost, diverse maze & $\alpha=1$; all frames \\
\bottomrule
\end{tabular}
\end{table}

\paragraph{GD and CEM planners.}
On PushObj and PointMaze, the planners use the settings of \cref{tab:impl_planner}.
Success is checked after every executed chunk.
The planning metric is $\norm{\bm{v}}_W^2=\text{MSE}_{\text{vis}}(\bm{v})+\alpha\,\text{MSE}_{\text{prop}}(\bm{v})$, the mean square of the 384 visual dimensions of $\bm{v}$ plus $\alpha$ times that of its 10 proprioceptive ones.
The planning cost applies it to the difference between predicted and goal latents, at the last predicted frame or at all frames (\cref{tab:impl_planner}).
PointMaze clips executed actions to $[-1,1]$.
The planner does not, and all learners, the baselines included, train on its actions before clipping.

\paragraph{DINO-WM.}
\label{sec:impl_robocasa}
On RoboCasa, we use the code of JEPA-WMs and its DINO-WM~\citep{zhou2025dinowm} checkpoint trained on DROID~\citep{khazatsky2024droid}, zero-shot.
CEM runs 15 iterations of 300 candidates with 10 elites over a horizon of three model steps, clips every sampled action dimension (position $\pm0.05$, rotation $\pm0.01$, gripper $\pm1$), and scores the last predicted visual latent by its L2 distance to the goal latent.
Each replan executes one 7-d model action, repeated for five raw steps, so an episode has at most 12 replans.
The model embeds each 7-d action, so the action token of a training window is the mean of its five raw actions.
It is the executed model action for a chunk-aligned window and a proportional mix of two actions for a dense window across two chunks.

\paragraph{LeWAM.}
\label{sec:impl_lewam}
On Puzzle and Scene, we use the code of LeWAM~\citep{fu2026lewam} and its released goal-reaching checkpoints, unchanged.
The model reads two history latents five raw steps apart ($k=5$, $L\le2$), the goal latent and the actions.
Its state stream regresses the next latent, and its action stream is a flow-matching policy.
At every replan, the policy proposes 32 action sequences that cover the steps left to the goal frame.
Five Adam steps (learning rate 0.05) refine each sequence to lower its cost, the distance between the latent predicted at its end and the goal latent.
A penalty of weight 0.01 keeps the refined sequence close to the proposal.
The first five actions of the lowest-cost sequence form the planned chunk $\va$.
Because the cost involves only the end of the sequence, a plan made with an adapted model can defer its correction for the shift to chunks that are never executed.
We therefore plan with the frozen model and add an inversion step.
From the same context, we replace $\va$ by the command $\bm{c}$ whose prediction under the method's adapted model is closest to the frozen model's prediction for $\va$.
We optimize $\bm{c}$ with 100 Adam steps (learning rate 0.2, normalized actions clipped to $\pm3$), starting from $\bm{c}=\va$.
All learners train on the commands they send, before the shift alters them.

\subsection{Baselines}
\label{sec:impl_base}

\paragraph{\adajepa.}
On PushObj and PointMaze, we run the released implementation in its default setting.
After every executed chunk, a new Adam optimizer~\citep{kingma2015adam} takes one step on the predictor's last transformer block and final LayerNorm (learning rate $5\times10^{-4}$) and on the encoder's projection head ($10^{-5}$).
The step uses at most nine chunk-aligned windows from the last five chunks.
The pretrained weights are restored before every episode.
\adajepa-P (\cref{sec:adajepa}) adapts only the predictor.

DINO-WM's encoder is frozen and has no projection head.
On DINO-WM, \adajepa therefore adapts only the predictor's last block and final LayerNorm, after every chunk but the last.
Its other settings are unchanged.

On LeWAM, the encoder also serves the policy and stays frozen.
\adajepa adapts, per environment, the layers of the last block that feed the state prediction and the final LayerNorm, in train mode.
After each chunk, it takes one Adam step ($5\times10^{-4}$ on the block) on the chunk-aligned windows of the five most recent chunks.
The same step also updates the action input layer, with a learning rate of $3\times10^{-2}$.
It is reset every episode.

\paragraph{\jepattt.}
We port the persistent test-time training rule of \jepattt~\citep{zhang2026jepattt} to our world models, planners and protocol, with the hyperparameters of its appendix.
Nothing was tuned on our tasks.
As in its paper, windows of four keyframes five raw steps apart, one starting at every raw step, fill a buffer.
AdamW (learning rate $5\times10^{-5}$, weight decay $10^{-3}$, batch 64, gradient clipping 1.0, modules in eval mode) minimizes the teacher-forced one-step latent loss on batches drawn uniformly from the buffer.
It updates once for every 64 new windows, at most once per replanning step.
After the first three chunks of an episode, each chunk adds up to five windows, so \jepattt updates about once every 13 chunks.
The buffer, weights and optimizer state are kept across episodes.
For a fair comparison, \jepattt runs with the same world models, planners and protocol as the other methods.
These are its only differences from its paper.
We train the action encoder and the whole predictor of the released \adajepa models instead of LeWorldModel's~\citep{maes2026lewm}.
We plan with GD and CEM toward a goal image instead of CEM on a reward head, and we run on our streams instead of 500-episode runs.
The variant \jepattt-EC updates after every chunk once the buffer holds 64 windows and is otherwise identical.

On DINO-WM, both variants keep these settings.
On LeWAM, they train all parameters outside the encoder on windows of three frames five raw steps apart, with the same optimizer and update schedules.

\subsection{\ours}
\label{sec:impl_sft}

\Cref{alg:sft} gives the deployment loop of \ours and \cref{tab:impl_sft} its hyperparameters.

\begin{algorithm}[!htb]
\caption{\ours during deployment}
\label{alg:sft}
\small
\begin{algorithmic}[1]
\Require pretrained predictor $\tz$; fast change $\theta_F\gets0$ (last block, final LN); LoRA $\theta_S\gets0$; buffer $\mathcal{D}\gets\emptyset$; Adam$_S$ on $\theta_S$; replan budget $N_{\max}$; step counts $K_F,K_S$; batch size $B$
\For{each episode}
  \For{$n=1,\dots,N_{\max}$}
    \State plan with $\tz+\theta_S+\theta_F$; execute the first chunk
    \State \textbf{if} goal reached \textbf{or} $n=N_{\max}$ \textbf{then break}
    \State $\mathcal{W}\gets$ dense windows of chunk $n$ $\cup$ replay windows
    \State Adam$_F\gets$ new Adam on $\theta_F$; take $K_F$ steps on $\ell(\mathcal{W})$
  \EndFor
  \State $\theta_F\gets0$ \Comment{forget this episode's correction}
  \State $\mathcal{D}\gets\mathcal{D}\,\cup$ dense windows of the episode
  \State $K_S$ Adam$_S$ steps, each on $\le B$ windows $\sim\mathcal{D}$, via $\tz+\theta_S$
\EndFor
\end{algorithmic}
\end{algorithm}

\begin{table}[!htb]
\centering
\caption{\textbf{\ours hyperparameters} for the PushObj and PointMaze world models, shared by all their tasks and planners (changes on DINO-WM and LeWAM in the text).
Both learners minimize the loss of \cref{eq:loss} with Adam ($\beta=(0.9,0.999)$, $\epsilon=10^{-8}$, no weight decay, no gradient clipping) in train mode (dropout 0.1), and planning runs in eval mode.}
\label{tab:impl_sft}
\footnotesize
\setlength{\tabcolsep}{3pt}
\begin{tabular}{@{}>{\raggedright\arraybackslash}p{0.2\linewidth}>{\raggedright\arraybackslash}p{0.37\linewidth}>{\raggedright\arraybackslash}p{0.37\linewidth}@{}}
\toprule
 & Fast learner & Slow learner \\
\midrule
Parameters & last predictor block and final LayerNorm (13 tensors, 3,314,848) & rank-8 LoRA, $\alpha=16$, on all 24 linear layers of the predictor (470,784); zero increment at start \\
Data & dense windows of the newest chunk
and one chunk-aligned window from each of the 8 previous chunks & dense windows of all episodes so far, in a buffer without size limit \\
Batch & all current windows
& $\min(B,|\mathcal{D}|)$ windows, $B=256$, drawn uniformly without replacement \\
Learning rate & $5\times10^{-4}$ & $5\times10^{-4}$ \\
Updates & $K_F=5$ steps after every chunk that does not end the episode & $K_S=2$ steps after every episode, after the fast reset \\
\bottomrule
\end{tabular}
\end{table}

\paragraph{Buffer and LoRA merging.}
The buffer stores the frame latents
(in half precision)
and actions of every episode, successful or not, and indexes all its dense windows, including those of the last chunk.
For planning and fast updates, the LoRA increment is merged into the linear weights, so planning costs the same as for \adajepa.

\paragraph{Other world models.}
On DINO-WM, the differences lie in the different model architectures and thus the number of learnable parameters are different.
The other settings follow \cref{tab:impl_sft}.
On LeWAM, the fast learner adapts the same parameters as \adajepa does there (\cref{sec:impl_base}).
A chunk has up to ten dense windows, and replay takes one random dense window from each of the three previous chunks.
The slow learner's LoRA covers the linear layers outside the encoder except the action and history input layers.
The other settings follow \cref{tab:impl_sft}.

\subsection{Evaluation protocol}
\label{sec:impl_protocol}

\paragraph{Runs and streams.}
On PushObj and PointMaze, each of the seeds 101, 102 and 103 fixes 50 start--goal pairs together with the environment and global random seeds, so every method plays the same 150 episodes per task (\cref{sec:setup}).
A method that resets every episode has one run per seed.
A method that learns across episodes plays the three 50-episode segments in sequence, carrying its learning state from one to the next, in the rotated orders 101$\to$102$\to$103, 102$\to$103$\to$101 and 103$\to$101$\to$102.
One such 150-episode stream is one run.
Within a stream, the environment and global random state of each segment follow its seed, whereas the learner's own random numbers (LoRA initialization, buffer sampling) follow the stream's first seed.

\paragraph{Stages and spread.}
Stage $k\in\{1,2,3\}$ comprises episodes $50k-49$ to $50k$ of every stream, averaged over the three orders.
The orders are rotations, so each stage covers every seed once, and stage 1 always starts from $\tz$.
Averages over groups of tasks are taken within each run before the mean and sample standard deviation (sd) across runs.

\Cref{fig:timescales}b and \cref{fig:stream} compare, in each stage, a method that learns across episodes with one that resets every episode.
In a stage, each order plays the 50 episodes of one seed.
We subtract the resetting method's success on that seed from the order's success in the stage.
The three orders give three differences, and we show their mean and sd.

\paragraph{RoboCasa, Puzzle and Scene.}
On RoboCasa, each task has 32 episodes, with the same start, goal and first frame for every method.
Every method is run with the seeds 1, 2 and 3.
A method that learns across episodes plays the 32 episodes of a task in order, as one stream per seed.
On Puzzle and Scene, each seed (42, 0 and 1) fixes the 50 start--goal pairs and the per-episode draws of the shift, the same for every method.
Methods that learn across episodes play them in sequence, as one stream per seed, starting from $\tz$.

\paragraph{Common random numbers.}
The controlled tests of \cref{sec:probe_planning} and the reruns marked in the captions of \cref{sec:supp_results,sec:supp_ablation} reset the random generators at the start of every episode.
The compared methods then start every episode with the same random numbers, so their differences come from the methods and not from the random draws.
The main experiments keep the released evaluation loop, in which one generator runs through all episodes of a run, and their sd across runs includes this variation.
The runs with common random numbers thus match those of \cref{tab:main} only in distribution and might have slightly different performances.

\section{More Details on the Error Analysis}
\label{sec:supp_probe}

This section details the measurements of \cref{sec:analysis}, all on PushObj with the CEM planner, shapes T and I and seeds 101--103 (150 episodes per shape before the exclusions of \cref{sec:probe_fan}), except the bars with the dataset's actions in \cref{fig:teaser}b.

\subsection{Action-fan probe}
\label{sec:probe_fan}

\paragraph{States and actions.}
The probe follows the frozen model's own episodes, run with common random numbers (\cref{sec:probe_planning}).
A probe point is the state after $n$ executed chunks, $n=0,\dots,N_i-1$ for an episode of $N_i$ chunks, recovered by replaying the executed actions from the initial state.
From each probe point, we execute 18 normalized 10-d chunks.
Nine form the fan $G$, namely the dataset's mean action, which is $\bm{0}$ for normalized actions, and four global pairs $\pm\va$ with $\va\sim\mathcal{N}(\bm{0},I)$.
The other nine, near the planner's choice, are the executed chunk $\va_{\text{e}}$ and four local pairs $\va_{\text{e}}\pm0.5\,\bm{\sigma}\odot\bm{\epsilon}$ with $\bm{\epsilon}\sim\mathcal{N}(\bm{0},I)$, where $\bm{\sigma}$ is the standard deviation of CEM's final distribution for that chunk.
Each chunk is executed from the replayed state, and the frame five raw steps later is encoded as its target.
Every model predicts this target in one step from the current frame alone ($L=1$), as in the planner's first rollout step.
\Cref{fig:teaser}b also shows all seven shapes with the dataset's actions, where no planner is involved.
For each shape, we replay 100 test segments of five chunks with their recorded actions and probe the state before each chunk (500 states per shape).

\paragraph{Metric and decomposition.}
The energies of \cref{eq:split} and the cosines of \cref{sec:analysis} and \cref{sec:probe_stats} use the PushObj planning metric (\cref{sec:impl_plan}, $\alpha=1$).
The nine chunks near the planner's choice do not enter the decomposition.
They check that an offset estimated on $G$ also holds at the actions the planner uses.
For the frozen model, subtracting $\ebar$ from their errors removes 61\% (T) and 57\% (I) of the error energy, against an offset share of 69\% and 79\% over $G$.

\paragraph{Ratios.}
All ratios compare a model $M$ with $\tz$ on the same states and actions.
The ratios $r_{\Cerr}$ and $r_{\Aerr}$ of \cref{fig:teaser}c, \cref{fig:plane} and \cref{tab:supp_decomp} divide the offset and action-dependent energy of $M$, summed over all probe points, by those of $\tz$.
\Cref{fig:timescales}a asks whether a correction helps in each episode, so it takes the offset ratio within one episode and averages it geometrically over episodes.
Halving the offset energy in one episode and doubling it in another then average to no change.

\paragraph{Methods that adapt within an episode.}
\adajepa, \adajepa-P, Fast only, Shift-Ada and Shift-Online update their model during an episode.
To probe them at the same states as the frozen model, we replay them along the frozen model's trajectory, chunk by chunk.
After each chunk, a method updates as it would in its own episode.
At the probe point after $n$ chunks of an episode, a method has made $n$ updates, and we probe the model it has then.
\adajepa also adapts the encoder's projection head, so \cref{tab:supp_decomp} scores it both in its own latent space and in $\tz$'s, in which all other models are scored.

\paragraph{Models trained on earlier episodes.}
Slow@100, \ours{}@100 and the two \jepattt{}@100 models are snapshots of the CEM streams of \cref{tab:main}, taken after 100 episodes.
A stream plays the three seeds in turn, 50 episodes each (\cref{sec:impl_protocol}), so such a snapshot has not trained on the third seed.
We freeze it and probe it on the frozen model's episodes of that seed.
A method's three snapshots, one per order, thus cover the same 150 episodes as every other model.
Slow@100 is the LoRA of a Slow-only stream, and \jepattt-EC@100 and \jepattt{}@100 are the weights of their streams.
\ours{}@100 w/o fast learner is the LoRA of an \ours stream.
\ours{}@100 adds a fast learner on top of this frozen LoRA, replayed as above and reset every episode.

\paragraph{Exclusions.}
A few start--goal pairs recur across seeds and a few segments overlap on the same test trajectory, so a snapshot may have trained on them.
We drop these episodes for all methods (17 on T, 13 on I).
This leaves 1,879 probe points in 133 episodes on T and 2,364 in 137 on I.

\subsection{Statistics of the offset}
\label{sec:probe_stats}

This subsection defines the offset statistics quoted in \cref{sec:offset,sec:specific,sec:dynamics}.
We write $\ebar_1,\dots,\ebar_N$ for the frozen model's offsets at the $N$ probe states of an episode and $\vb$ for their mean.
All cosines use the inner product $\langle\bm{u},\bm{v}\rangle_W$ of the planning metric, for which $\langle\bm{v},\bm{v}\rangle_W=\norm{\bm{v}}_W^2$.

\paragraph{Robustness of the offset share.}
With the planner's actions, the offset share of $\tz$ is 0.69 on T and 0.79 on I (\cref{fig:teaser}b).
It changes little under other choices.
With two or three context frames, it is 0.71--0.72 on T and 0.80--0.81 on I.
Without the exclusions of \cref{sec:probe_fan}, it is 0.68 and 0.79.
On the states that \adajepa, Fast only and \ours{}@100 visit in their own episodes, it is 0.76--0.79 (I, seed 103).

\paragraph{Shifted goal.}
As mentioned in \cref{sec:offset}, an uncorrected offset steers the true state toward $\vg-\vb$ rather than $\vg$.
A successful episode ends at the goal, so this can show only in failed episodes.
We therefore take the failed episodes of the frozen model (69 on T, 103 on I).
The final deviation of an episode is its last latent minus the goal latent.
If the goal is shifted, this deviation points along the negative of the episode's mean offset $\vb$.
The cosine between the two, averaged over these episodes, is 0.41 on T and 0.22 on I.
With the mean offset of another episode of the same shape and seed, it is about 0.
The deviation thus follows the episode's own offset, though only in part.

\paragraph{Drift within an episode.}
Within an episode, the offset drifts slowly (\cref{sec:specific}).
We measure this by the cosine between the offsets of two probe states of the same episode that lie $j$ chunks apart,
\begin{equation}
  \cos_j=\frac{\sum\langle\ebar_n,\ebar_{n+j}\rangle_W}{\sqrt{\sum\norm{\ebar_n}_W^2\,\sum\norm{\ebar_{n+j}}_W^2}},
  \label{eq:lagcos}
\end{equation}
where each sum runs over all such pairs in all episodes.
Pairs with larger offsets thus weigh more.
At a gap of one chunk, the cosine is 0.87 on T and 0.89 on I (1,746 and 2,227 pairs).
At a gap of 12 chunks, it is still 0.57 and 0.55 (584 and 838 pairs).

\paragraph{Carried-over corrections.}
A correction carried over from the previous episode is nearly orthogonal to the one the new episode needs (\cref{sec:specific}).
Let $M$ be the frozen model with the previous episode's shift or fast weights.
At each probe state of the new episode, $M$ changes the offset by $\ebar_M-\ebar_{\tz}$.
The change that would remove the offset is $-\ebar_{\tz}$.
We take the cosine of the two, pooled over all probe states as in \cref{eq:lagcos}.
It is 0.05 on T and 0.09 on I for the Shift-Prev vector and 0.05 and 0.10 for the fast learner's weights.
Such a correction removes little of the new offset and adds its own energy, so the offset energy rises (\cref{fig:timescales}a).

\paragraph{Offset and state.}
\Cref{sec:dynamics} compares the offsets of two probe states from different episodes of a shape.
We describe a state by five coordinates, the agent position, the block position and the block angle.
Each coordinate is scaled by its standard deviation over the shape's probe states, and the state distance is the Euclidean distance between two such states.
We call a pair close if this distance is below 0.5 and distant if it is 1 or more.
The cosine of the offsets is pooled over all pairs of a group as in \cref{eq:lagcos}.
It is 0.57 on T and 0.71 on I for close pairs (2,631 and 4,807 pairs), 0.30 and 0.36 for pairs in between, and 0.06 and 0.12 for distant ones.
Close states thus have similar offsets even in different episodes.
On PushObj, however, close pairs make up only 0.15\% (T) and 0.17\% (I) of the pairs from different episodes, so an episode rarely comes close to the states of another.

\subsection{Full error decomposition}
\label{sec:supp_decomp}

\begin{table}[!htb]
\centering
\caption{\textbf{What each model changes in the prediction error} (PushObj, CEM).
Offset and action-dependent energy ($r_{\Cerr}$, $r_{\Aerr}$) relative to $\tz$.
Shifts change only the offset, so their $r_{\Aerr}$ is 1.00.
These ratios and the @100 models are defined in \cref{sec:probe_fan}.}
\label{tab:supp_decomp}
\scriptsize
\begin{tabular*}{0.6\textwidth}{@{\extracolsep{\fill}}l cccc@{}}
\toprule
 & \multicolumn{2}{c}{T (seen)} & \multicolumn{2}{c}{I (unseen)} \\
\cmidrule(lr){2-3} \cmidrule(l){4-5}
Model & $r_{\Cerr}$ & $r_{\Aerr}$ & $r_{\Cerr}$ & $r_{\Aerr}$ \\
\midrule
\multicolumn{5}{@{}l}{\emph{Within one episode, reset after every episode}} \\
\adajepa & 0.49 & 0.96 & 0.46 & 0.93 \\
\adajepa, in $\tz$'s latent space & 0.53 & 0.97 & 0.49 & 0.96 \\
\adajepa-P & 0.51 & 0.97 & 0.47 & 0.95 \\
Shift-Ada & 0.44 & 1.00 & 0.51 & 1.00 \\
Shift-Online & 0.30 & 1.00 & 0.39 & 1.00 \\
Fast only & 0.33 & 0.95 & 0.28 & 0.92 \\
\midrule
\multicolumn{5}{@{}l}{\emph{Carried over from the previous episode}} \\
Shift-Prev & 2.19 & 1.00 & 1.97 & 1.00 \\
\midrule
\multicolumn{5}{@{}l}{\emph{Trained on 100 earlier episodes, probed on episodes not trained on}} \\
Slow@100 & 0.70 & 0.96 & 0.42 & 0.82 \\
\ours{}@100 w/o fast learner & 0.74 & 0.98 & 0.44 & 0.82 \\
\ours{}@100 & 0.30 & 0.94 & 0.19 & 0.81 \\
\jepattt-EC@100 & 0.72 & 0.82 & 0.33 & 0.65 \\
\jepattt{}@100 & 0.80 & 0.97 & 0.53 & 0.89 \\
\bottomrule
\end{tabular*}
\end{table}

\begin{figure*}[!htb]
  \centering
  \includegraphics[width=\textwidth]{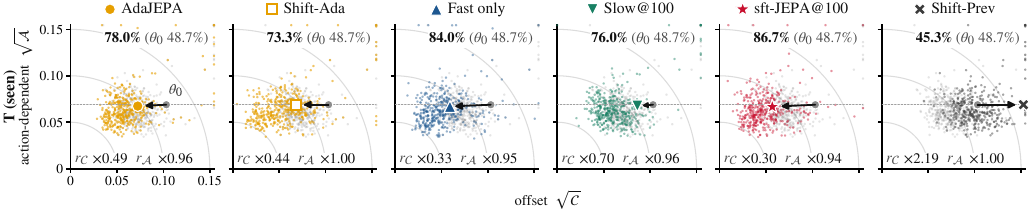}
  \caption{\textbf{What each method fixes on the seen PushObj shape T}, drawn as in \cref{fig:plane}.}
  \label{fig:supp_plane}
\end{figure*}

\Cref{tab:controlled} compares the models of \cref{sec:analysis} by planning success.
\Cref{tab:supp_decomp} complements it with their prediction errors, split into offset and action-dependent energy.
As in \cref{fig:plane}, each ratio divides a model's energy, summed over all probe points, by that of $\tz$.
\Cref{fig:timescales}a asks whether a correction helps in each episode, so it averages per-episode ratios (\cref{sec:probe_fan}).
Thus, the two can differ, \eg, $\times1.97$ and $\times1.86$ for Shift-Prev on I.
\jepattt-EC@100 trains the whole predictor and lowers the action-dependent error most ($\times0.65$ on I).
However, it leaves more of the offset than \ours{}@100 ($\times0.33$ against $\times0.19$).
This agrees with \cref{sec:supp_tttfast}, where adding our fast learner to \jepattt-EC raises its success rate.

\Cref{fig:supp_plane} complements \cref{fig:plane} with the seen shape T.
On T, the action-dependent error falls little for all models shown ($\times0.94$--$1.00$).
The frozen model may already predict the effects of actions well on this seen shape.
In both figures, each large marker is the root mean square of its model's dots, the dotted line passes through that of $\tz$, and the arcs mark equal total error.
The ratio $r_{\Aerr}$ is in bold when it is below $\times0.9$, and states beyond the axes are drawn on the frame.

\subsection{Controlled planning tests}
\label{sec:probe_planning}

\paragraph{Same episodes and random numbers.}
Every model of \cref{tab:controlled} plans the same 150 episodes per shape.
A @100 snapshot plans the 50 episodes of the seed it has not trained on, so a method's three snapshots cover the same 150 episodes (\cref{sec:probe_fan}).
All models plan with common random numbers (\cref{sec:impl_protocol}).
At the start of each episode, the CPU and GPU random generators are reset to $1000\cdot\text{seed}+\text{episode index}$.
CEM draws only from the CPU generator, so an episode receives the same CEM noise with every model.
As a check, every model that starts from $\tz$ executed the same first chunk as Frozen $\tz$ in all 50 episodes of each shape and seed.
Frozen $\tz$, \adajepa and Fast only were rerun this way as well, so their numbers match those of the main experiments only in distribution.

\paragraph{Shifts.}
A shift is a vector $\bm{s}\in\mathbb{R}^{394}$ that we write into the bias of the predictor's final LayerNorm, on the visual and proprioceptive dimensions.
It is thus added to every candidate's prediction at every rollout step and propagates through the autoregressive context, while the encodings of the current frame and the goal stay unchanged.
For Shift-Ada and Shift-Online, $\bm{s}$ is the negative of $\tz$'s mean error on a set of windows, so adding it subtracts this error as in \cref{sec:specific}.
Shift-Ada uses \adajepa's at most nine chunk-aligned windows, and Shift-Online the fast learner's dense and replay windows.
Both are refitted from scratch after every chunk and start each episode at $\bm{s}=\bm{0}$.
Shift-Prev instead keeps the previous episode's final Shift-Online vector for the whole episode ($\bm{s}=\bm{0}$ in the first episode of a run).
For \adajepa-P's shift $\Delta$, we take the mean change of its prediction from that of $\tz$ at the current frame over 33 symmetric actions ($\bm{0}$ and 16 pairs $\pm\va$).
The average keeps the part of the change that all actions share.
The actions are only fed to the two models and are not executed, so $\Delta$ needs no true outcome.
\adajepa-P without its shift plans with $\Delta$ subtracted from the same bias.
We undo this subtraction before \adajepa-P's next update, so its adaptation is not affected.
For $\tz$ + \adajepa-P's shift, a copy of \adajepa-P adapts as usual but does not plan.
The planner uses $\tz$ with this copy's $\Delta$ added to the bias.

\paragraph{Longer episodes.}
The dashed line in \cref{fig:timescales}b is a variant on T with longer episodes.
Both the never-reset fast learner and the Fast only it is compared with are allowed 30 replans, and success is counted within 30.
The zero line thus stands for Fast only with the same number of replans as each curve.
With more updates per episode, the loss from never resetting grows over the stream, to 10.7, 18.0 and 24.0 points in the three stages.

\section{More Results}
\label{sec:supp_results}

\subsection{Detailed results of \texorpdfstring{\Cref{tab:main}}{Table 2}}
\label{sec:supp_bench}

\begin{table*}[!htb]
\centering
\caption{\textbf{Planning success rate (\%) on PushObj and PointMaze}.
The detailed per-task version of \cref{tab:main} in the main paper. $^{\dagger}$Learns across episodes.}
\label{tab:supp_pertask}
\scriptsize
\setlength{\tabcolsep}{4pt}
\begin{tabular}{@{}ll ccccccc@{}}
\toprule
 & Task & Frozen $\tz$ & \adajepa & \jepattt{}$^{\dagger}$ & \jepattt-EC{}$^{\dagger}$ & Fast only & Slow only{}$^{\dagger}$ & \ours{}$^{\dagger}$ \\
\midrule
\multicolumn{9}{@{}l}{\emph{GD planner}} \\
\multirow{4}{*}{PointMaze} & Default & 86.0\std{8.7} & 86.7\std{7.6} & 86.7\std{1.8} & 84.4\std{0.8} & 85.3\std{5.0} & \textbf{87.3}\std{2.4} & 85.6\std{1.5} \\
 & Low mass & 82.7\std{2.3} & \textbf{84.0}\std{7.2} & 80.9\std{3.4} & 79.1\std{2.8} & 82.0\std{5.3} & 82.4\std{1.4} & \textbf{84.0}\std{0.7} \\
 & High damping & 75.3\std{4.2} & 80.0\std{2.0} & 78.9\std{1.4} & 81.1\std{1.0} & 82.7\std{3.1} & 84.0\std{2.7} & \textbf{85.1}\std{0.8} \\
 & Unseen layout & 60.7\std{2.3} & 65.3\std{6.4} & 65.1\std{1.4} & \textbf{80.7}\std{3.1} & 67.3\std{3.1} & 70.0\std{3.7} & 77.3\std{2.9} \\
\cmidrule(l){2-9}
\multirow{4}{*}{PushObj, seen} & T & 51.3\std{7.6} & 79.3\std{5.0} & 67.8\std{4.0} & 74.2\std{3.3} & 80.7\std{4.2} & 63.6\std{3.0} & \textbf{85.1}\std{1.4} \\
 & L & 50.7\std{1.2} & 82.7\std{5.0} & 65.1\std{3.4} & 71.6\std{4.3} & 82.7\std{3.1} & 66.4\std{4.7} & \textbf{83.8}\std{1.7} \\
 & Z & 45.3\std{1.2} & \textbf{84.7}\std{4.2} & 66.0\std{1.3} & 70.0\std{4.2} & 84.0\std{3.5} & 64.4\std{1.9} & 82.9\std{1.5} \\
 & + & 40.0\std{8.7} & 77.3\std{3.1} & 62.4\std{3.8} & 68.4\std{3.0} & \textbf{80.7}\std{2.3} & 61.1\std{2.8} & 79.8\std{1.0} \\
\cmidrule(l){2-9}
\multirow{3}{*}{PushObj, unseen} & I & 27.3\std{4.6} & 39.3\std{4.2} & 40.4\std{2.7} & 53.1\std{2.8} & 37.3\std{3.1} & 44.2\std{2.7} & \textbf{66.9}\std{4.3} \\
 & Small T & 44.0\std{6.9} & 67.3\std{3.1} & 62.4\std{2.5} & 64.2\std{4.4} & 74.7\std{3.1} & 63.6\std{3.4} & \textbf{76.9}\std{3.4} \\
 & Square & 19.3\std{3.1} & 43.3\std{6.4} & 45.8\std{1.7} & 56.9\std{3.0} & 42.0\std{3.5} & 50.2\std{0.8} & \textbf{64.2}\std{2.1} \\
\midrule
\multicolumn{9}{@{}l}{\emph{CEM planner}} \\
\multirow{4}{*}{PointMaze} & Default & 83.3\std{4.2} & 84.7\std{5.0} & \textbf{85.3}\std{0.7} & 83.3\std{1.8} & 84.0\std{5.3} & 83.6\std{1.4} & \textbf{85.3}\std{1.2} \\
 & Low mass & 83.3\std{5.0} & 85.3\std{3.1} & 85.1\std{1.7} & 86.2\std{1.4} & 87.3\std{4.2} & \textbf{88.0}\std{1.2} & 87.6\std{1.5} \\
 & High damping & 76.0\std{3.5} & 77.3\std{1.2} & 79.6\std{3.7} & 80.4\std{1.4} & 82.7\std{2.3} & 81.8\std{2.0} & \textbf{83.6}\std{1.7} \\
 & Unseen layout & 54.0\std{2.0} & 56.7\std{5.0} & 58.9\std{3.8} & 74.2\std{1.0} & 57.3\std{6.4} & 70.9\std{4.1} & \textbf{74.7}\std{2.3} \\
\cmidrule(l){2-9}
\multirow{4}{*}{PushObj, seen} & T & 48.0\std{10.6} & 79.3\std{5.0} & 68.7\std{0.7} & 70.9\std{4.3} & \textbf{86.0}\std{4.0} & 71.1\std{1.4} & 85.3\std{1.3} \\
 & L & 53.3\std{5.8} & 77.3\std{1.2} & 64.7\std{3.5} & 68.2\std{1.7} & 83.3\std{2.3} & 64.9\std{0.4} & \textbf{85.3}\std{2.4} \\
 & Z & 37.3\std{8.1} & 72.0\std{9.2} & 59.6\std{3.3} & 66.9\std{3.7} & 82.7\std{4.2} & 56.4\std{2.5} & \textbf{83.6}\std{1.0} \\
 & + & 42.0\std{3.5} & 68.0\std{4.0} & 56.7\std{2.7} & 60.9\std{5.6} & 78.0\std{3.5} & 56.2\std{1.7} & \textbf{80.4}\std{2.1} \\
\cmidrule(l){2-9}
\multirow{3}{*}{PushObj, unseen} & I & 24.0\std{6.9} & 48.7\std{10.3} & 47.8\std{2.5} & 52.9\std{4.4} & 46.7\std{10.3} & 50.9\std{1.4} & \textbf{70.0}\std{2.9} \\
 & Small T & 47.3\std{8.1} & 67.3\std{6.4} & 62.9\std{2.5} & 63.1\std{2.5} & 71.3\std{9.0} & 62.9\std{2.1} & \textbf{77.8}\std{4.5} \\
 & Square & 22.7\std{5.0} & 42.7\std{3.1} & 44.0\std{3.5} & 58.2\std{2.8} & 49.3\std{3.1} & 52.7\std{4.4} & \textbf{64.2}\std{5.4} \\
\bottomrule
\end{tabular}
\end{table*}

\Cref{tab:main} reports success rates averaged over groups of tasks.
\Cref{tab:supp_pertask} gives them for each of the seven PushObj shapes and the four PointMaze settings, under both planners.

\subsection{\adajepa with more steps and our slow learner}
\label{sec:supp_adasteps}

\begin{table*}[!htb]
\centering
\caption{\textbf{\adajepa with more learning steps and our slow learner} (success rate (\%), columns as in \cref{tab:main}).
Steps are Adam steps per chunk of \adajepa or the fast learner.
\adajepa's runs use common random numbers (\cref{sec:impl_protocol}).
$^{\dagger}$Learns across episodes.}
\label{tab:supp_adasteps}
\scriptsize
\setlength{\tabcolsep}{2.2pt}
\begin{tabular*}{\textwidth}{@{\extracolsep{\fill}}ll cccccc cccccc@{}}
\toprule
 & & \multicolumn{6}{c}{PushObj} & \multicolumn{6}{c}{PointMaze} \\
\cmidrule(lr){3-8} \cmidrule(l){9-14}
 & & \multicolumn{3}{c}{GD} & \multicolumn{3}{c}{CEM} & \multicolumn{3}{c}{GD} & \multicolumn{3}{c}{CEM} \\
\cmidrule(lr){3-5} \cmidrule(lr){6-8} \cmidrule(lr){9-11} \cmidrule(l){12-14}
Method & Steps & Seen & Unseen & All & Seen & Unseen & All & Medium & Layout & All & Medium & Layout & All \\
\midrule
\multicolumn{14}{@{}l}{\emph{Reset every episode}} \\
\adajepa & 1 & 80.7\std{3.8} & 51.1\std{1.4} & 68.0\std{2.3} & 75.5\std{2.8} & 54.9\std{1.0} & 66.7\std{2.0} & 83.6\std{4.0} & 61.3\std{6.4} & 78.0\std{4.4} & 82.2\std{2.5} & 55.3\std{5.0} & 75.5\std{3.0} \\
 & 3 & 81.5\std{2.2} & 56.2\std{2.1} & 70.7\std{2.1} & 80.7\std{2.0} & 56.9\std{3.0} & 70.5\std{1.4} & 82.9\std{4.5} & 68.0\std{5.3} & 79.2\std{4.5} & 84.0\std{4.7} & 58.7\std{6.4} & 77.7\std{4.5} \\
 & 5 & 83.5\std{2.3} & 53.3\std{4.8} & 70.6\std{3.0} & 83.3\std{3.8} & 60.7\std{4.0} & 73.6\std{3.8} & 83.3\std{4.0} & 73.3\std{4.2} & 80.8\std{3.4} & 84.7\std{3.7} & 58.0\std{7.2} & 78.0\std{4.6} \\
 & 7 & 82.2\std{5.1} & 52.7\std{4.7} & 69.5\std{4.7} & 82.7\std{1.8} & 58.7\std{0.0} & 72.4\std{1.0} & 82.7\std{5.0} & 69.3\std{5.0} & 79.3\std{4.5} & 84.0\std{3.7} & 67.3\std{5.8} & 79.8\std{4.0} \\
Fast only & 5 & 82.0\std{0.9} & 51.3\std{1.2} & 68.9\std{0.5} & 82.5\std{0.5} & 55.8\std{5.7} & 71.0\std{2.3} & 83.3\std{3.7} & 67.3\std{3.1} & 79.3\std{2.1} & 84.7\std{1.8} & 57.3\std{6.4} & 77.8\std{2.9} \\
\midrule
\multicolumn{14}{@{}l}{\emph{With our slow learner (150-episode streams)}} \\
\adajepa{} + slow learner$^{\dagger}$ & 5 & 83.0\std{1.2} & 67.7\std{1.7} & 76.4\std{1.0} & 83.2\std{2.0} & 69.8\std{0.4} & 77.5\std{1.3} & 84.1\std{1.3} & 78.4\std{3.4} & 82.7\std{1.7} & 84.9\std{0.8} & 72.2\std{1.7} & 81.7\std{0.9} \\
\ours{}$^{\dagger}$ & 5 & 82.9\std{0.5} & 69.3\std{1.0} & 77.1\std{0.6} & 83.7\std{0.6} & 70.7\std{2.9} & 78.1\std{0.9} & 84.9\std{0.2} & 77.3\std{2.9} & 83.0\std{0.6} & 85.5\std{0.7} & 74.7\std{2.3} & 82.8\std{0.7} \\
\bottomrule
\end{tabular*}
\end{table*}

In the main experiments, \adajepa runs in its default setting, with one Adam step per chunk (\cref{sec:impl_base}).
Here, we further examine how it performs with more steps. From \cref{tab:supp_adasteps}, more steps help \adajepa to some extent.
From one to five steps, its overall success rises by 2.6 (GD) and 6.9 points (CEM) on PushObj and by 2.8 and 2.5 on PointMaze.
Beyond five steps, success rate drops again in three of the four settings, by 1.1--1.5 points at seven steps.
Even with its best step count in each column, \adajepa still underperforms \ours a lot.
The gap is largest on unseen shapes (10.0--13.1) and on the held-out layouts (4.0--7.4).
We also add our slow learner to five-step \adajepa.
The slow learner raises success by 14.4 (GD) and 9.1 points (CEM) on unseen shapes and by 5.1 and 14.2 on the held-out layouts.
These gains further validate the effectiveness of our slow-fast learning mechanism.

\subsection{\jepattt with a larger learning rate and our fast learner}
\label{sec:supp_tttfast}

\begin{table*}[!htb]
\centering
\caption{\textbf{\jepattt with a larger learning rate and our fast learner} (success rate (\%), columns as in \cref{tab:main}).
All methods learn across episodes on the 150-episode streams.
LR is the learning rate of \jepattt's learner.
\jepattt-EC + fast learner adds our fast learner, reset every episode, to \jepattt-EC.
}
\label{tab:supp_ttt}
\scriptsize
\setlength{\tabcolsep}{1.9pt}
\begin{tabular*}{\textwidth}{@{\extracolsep{\fill}}ll cccccc cccccc@{}}
\toprule
 & & \multicolumn{6}{c}{PushObj} & \multicolumn{6}{c}{PointMaze} \\
\cmidrule(lr){3-8} \cmidrule(l){9-14}
 & & \multicolumn{3}{c}{GD} & \multicolumn{3}{c}{CEM} & \multicolumn{3}{c}{GD} & \multicolumn{3}{c}{CEM} \\
\cmidrule(lr){3-5} \cmidrule(lr){6-8} \cmidrule(lr){9-11} \cmidrule(l){12-14}
Method & LR & Seen & Unseen & All & Seen & Unseen & All & Medium & Layout & All & Medium & Layout & All \\
\midrule
\multicolumn{14}{@{}l}{\emph{Learning rate}} \\
\jepattt & $5\times10^{-5}$ & 65.3\std{2.7} & 49.6\std{1.0} & 58.6\std{1.7} & 62.4\std{2.0} & 51.6\std{1.1} & 57.7\std{0.7} & 82.1\std{1.2} & 65.1\std{1.4} & 77.9\std{0.9} & 83.3\std{0.8} & 58.9\std{3.8} & 77.2\std{1.0} \\
 & $5\times10^{-4}$ & 60.9\std{0.4} & 50.7\std{1.9} & 56.6\std{1.0} & 58.6\std{2.1} & 53.0\std{2.0} & 56.2\std{2.0} & 81.9\std{1.5} & 73.3\std{2.0} & 79.8\std{0.9} & 82.1\std{0.1} & 66.4\std{5.7} & 78.2\std{1.4} \\
\jepattt-EC & $5\times10^{-5}$ & 71.1\std{3.4} & 58.1\std{1.6} & 65.5\std{2.6} & 66.7\std{1.5} & 58.1\std{2.6} & 63.0\std{1.9} & 81.6\std{0.8} & 80.7\std{3.1} & 81.3\std{0.7} & 83.3\std{0.8} & 74.2\std{1.0} & 81.1\std{0.8} \\
 & $5\times10^{-4}$ & 67.6\std{2.1} & 61.3\std{1.6} & 64.9\std{1.5} & 65.4\std{2.8} & 60.7\std{3.6} & 63.4\std{3.0} & 85.0\std{0.3} & 80.7\std{7.2} & 83.9\std{1.8} & 84.2\std{0.6} & 80.7\std{2.0} & 83.3\std{0.2} \\
\midrule
\multicolumn{14}{@{}l}{\emph{With our fast learner, reset every episode}} \\
\jepattt-EC + fast learner & $5\times10^{-5}$ & 82.9\std{0.3} & 69.9\std{1.9} & 77.4\std{0.9} & 84.6\std{1.4} & 70.9\std{1.4} & 78.7\std{0.5} & 83.5\std{2.0} & 80.9\std{1.7} & 82.8\std{1.3} & 84.4\std{1.3} & 80.2\std{1.4} & 83.3\std{1.3} \\
\multicolumn{2}{@{}l}{\ours} & 82.9\std{0.5} & 69.3\std{1.0} & 77.1\std{0.6} & 83.7\std{0.6} & 70.7\std{2.9} & 78.1\std{0.9} & 84.9\std{0.2} & 77.3\std{2.9} & 83.0\std{0.6} & 85.5\std{0.7} & 74.7\std{2.3} & 82.8\std{0.7} \\
\bottomrule
\end{tabular*}
\end{table*}

In the main experiments, both \jepattt variants follow the setting of their paper (\cref{sec:impl_base}).
They update the action encoder and the whole predictor with a learning rate of $5\times10^{-5}$ and keep the weights across episodes.
Here, we further examine how they perform with a larger learning rate $5\times10^{-4}$.
From \cref{tab:supp_ttt}, the larger learning rate helps on PointMaze, where the overall success rate rises by 1.0--2.6 points.
On PushObj, it changes the overall success rate by only $-2.0$ to $+0.4$ points.
With either learning rate, both variants still underperform \ours a lot on PushObj, by 11.6--21.9 points.
We also add our fast learner to \jepattt-EC.
The fast learner raises its overall success rate by 11.9 (GD) and 15.7 points (CEM) on PushObj and by 1.5 and 2.2 on PointMaze.
These gains again validate our slow-fast learning mechanism.

Finally, our slow learner updates only after an episode ends, whereas \jepattt-EC updates within the episode.
A minor benefit is that the learning time of the slow learner falls between episodes rather than within them.
This is a small design choice, not our core contribution.
Our core contribution is the analysis of what adaptation corrects in latent world models and the slow-fast learning mechanism that follows from it.
One can also spread the slow updates over the chunks of an episode.

\subsection{Larger learning rates on the held-out layouts}
\label{sec:supp_lr}

\begin{table*}[!htb]
\centering
\caption{\textbf{Success rate (\%) on PointMaze held-out layouts under larger learning rates}.
The learning rate applies to \adajepa's predictor, our Fast only learner, and \ours.
Indented rows show \ours in the three 50-episode stages of its streams.
$^{\dagger}$Learns across episodes.}
\label{tab:supp_lr}
\scriptsize
\begin{tabular*}{0.82955\textwidth}{@{\extracolsep{\fill}}l ccc ccc@{}}
\toprule
 & \multicolumn{3}{c}{GD} & \multicolumn{3}{c}{CEM} \\
\cmidrule(lr){2-4} \cmidrule(l){5-7}
Learning rate & $5\times10^{-4}$ & $10^{-3}$ & $5\times10^{-3}$ & $5\times10^{-4}$ & $10^{-3}$ & $5\times10^{-3}$ \\
\midrule
\adajepa & 65.3\std{6.4} & 66.7\std{11.0} & 75.3\std{1.2} & 56.7\std{5.0} & 62.0\std{9.2} & 61.3\std{8.3} \\
Fast only & 67.3\std{3.1} & 69.3\std{1.2} & 70.0\std{3.5} & 57.3\std{6.4} & 60.7\std{7.6} & 62.7\std{5.0} \\
\ours{}$^{\dagger}$ & 77.3\std{2.9} & 80.4\std{3.4} & 82.0\std{3.5} & 74.7\std{2.3} & 76.4\std{1.9} & 75.6\std{1.5} \\
\quad Episodes 1--50 & 67.3\std{1.2} & 72.0\std{2.0} & 74.7\std{2.3} & 59.3\std{3.1} & 64.7\std{6.4} & 61.3\std{3.1} \\
\quad Episodes 51--100 & 76.7\std{4.2} & 80.7\std{9.5} & 81.3\std{13.3} & 78.0\std{7.2} & 78.0\std{9.2} & 78.7\std{4.6} \\
\quad Episodes 101--150 & 88.0\std{3.5} & 88.7\std{2.3} & 90.0\std{2.0} & 86.7\std{3.1} & 86.7\std{7.0} & 86.7\std{3.1} \\
\bottomrule
\end{tabular*}
\end{table*}

In \cref{tab:main}, \ours is slightly below \jepattt-EC on the held-out layouts with GD.
As mentioned in \cref{sec:results}, each layout recurs about 30 times per stream, so the more frequent updates of \jepattt-EC may give it an advantage.
\ours may thus learn too little in this scenario.
Here, we further examine whether faster learning within the episode can compensate for this.
We raise the learning rate shared by \adajepa and our fast learner to \{$10^{-3}$, $5\times10^{-3}$\}.
From \cref{tab:supp_lr}, the larger rates help \adajepa, Fast only and \ours.
With GD, \ours rises to 80.4\% and 82.0\%, matching or exceeding \jepattt-EC.
Its gain is largest in the first 50 episodes, when the slow learner has seen each layout only a few times, and shrinks afterward.
\adajepa still underperforms \ours at every rate, by 6.7--13.7 (GD) and 14.3--18.0 points (CEM).
Note that these runs are only for observation.
The main experiments keep one learning rate for \adajepa and our learners on all tasks, for a consistent comparison without specific tuning on a given task.

\section{More Ablations and Discussions}
\label{sec:supp_ablation}

\begin{figure}[!htb]
  \centering
  \includegraphics[width=0.55\textwidth]{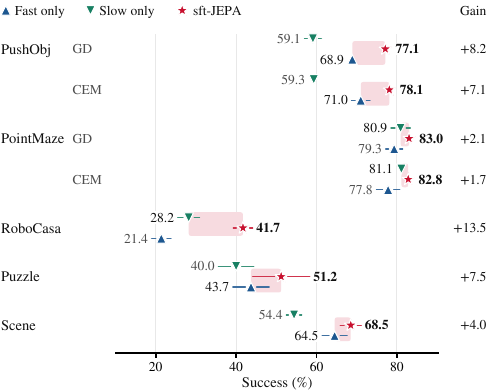}
  \caption{\textbf{Success rates (\%) of each learner alone and both together.}
  The numbers are as in the All and Avg columns of \cref{tab:main,tab:other}.
  Markers are shifted vertically for clarity.
  Band and gain show by how many points \ours leads the better learner alone.}
  \label{fig:ablation}
\end{figure}

\paragraph{Effects of slow and fast learner alone.}
\Cref{fig:ablation} provides a more intuitive view of the comparison between each learner alone and \ours in \cref{tab:main,tab:other}.
In all seven settings, \ours leads the better of the two learners, so the two learners complement each other.

\begin{table*}[!htb]
\centering
\caption{\textbf{Effects of dense windows and learning steps in the fast learner} (success rate (\%) on PushObj).
Fast only is trained on \adajepa's windows or on our dense windows (with replay windows), with one and five Adam steps per chunk.
The last row is the setting of our fast learner.
All runs use common random numbers (\cref{sec:impl_protocol}), so the last row differs slightly from Fast only in \cref{tab:main}.}
\label{tab:supp_fast}
\scriptsize
\begin{tabular*}{0.82955\textwidth}{@{\extracolsep{\fill}}ll cccccc@{}}
\toprule
 & & \multicolumn{3}{c}{GD} & \multicolumn{3}{c}{CEM} \\
\cmidrule(lr){3-5} \cmidrule(l){6-8}
Windows & Steps & Seen & Unseen & All & Seen & Unseen & All \\
\midrule
\adajepa's windows & 1 & 81.2\std{2.5} & 50.9\std{5.4} & 68.2\std{3.3} & 74.5\std{2.2} & 52.9\std{2.8} & 65.2\std{2.0} \\
 & 5 & 81.2\std{0.3} & 48.7\std{5.7} & 67.2\std{2.3} & 82.2\std{1.2} & 55.8\std{4.7} & 70.9\std{2.2} \\
Dense windows & 1 & 80.2\std{4.5} & 51.6\std{2.3} & 67.9\std{3.2} & 77.2\std{3.3} & 53.3\std{3.5} & 67.0\std{3.2} \\
 & 5 & 82.2\std{2.1} & 50.2\std{5.8} & 68.5\std{3.6} & 82.3\std{3.1} & 57.6\std{5.1} & 71.7\std{3.7} \\
\bottomrule
\end{tabular*}
\end{table*}

\paragraph{Effects of dense windows and learning steps in the fast learner.}
Our fast learner adapts only the predictor's last block and final LayerNorm.
To let it learn more within each episode, we give it five Adam steps per chunk and train it on dense windows.
Here, we use Fast only to examine how much these two choices contribute (\cref{tab:supp_fast}).
On average over GD and CEM, five steps instead of one raise the overall success rate by 2.4 points ($-1.0$ with GD, $+5.7$ with CEM).
Dense windows add another 1.1 points ($+1.3$ and $+0.8$).
Overall, these two design tweaks yield only modest gains.
The main gains of \ours come from its slow-fast learning mechanism.
For example, adding the slow learner to Fast only raises the overall success rate by 8.2 (GD) and 7.1 points (CEM) in \cref{tab:main}, and by 18.0 and 14.9 points on unseen shapes.

\begin{table*}[!htb]
\centering
\caption{\textbf{More results \wrt effects of resetting the fast learner} (success rate (\%) on PushObj and PointMaze).
The detailed per-task version of \cref{tab:reset}, extended to PointMaze.
Reset: the fast learner restarts every episode from $\tz$.
Never: it keeps its change over the 150-episode streams.
}
\label{tab:supp_noreset}
\scriptsize
\setlength{\tabcolsep}{3.9pt}
\begin{tabular*}{0.82955\textwidth}{@{\extracolsep{\fill}}ll cccc cccc@{}}
\toprule
 & & \multicolumn{4}{c}{GD} & \multicolumn{4}{c}{CEM} \\
\cmidrule(lr){3-6} \cmidrule(l){7-10}
 & & \multicolumn{2}{c}{Fast only} & \multicolumn{2}{c}{\ours{}} & \multicolumn{2}{c}{Fast only} & \multicolumn{2}{c}{\ours{}} \\
\cmidrule(lr){3-4} \cmidrule(lr){5-6} \cmidrule(lr){7-8} \cmidrule(l){9-10}
 & Task & Reset & Never & Reset & Never & Reset & Never & Reset & Never \\
\midrule
\multirow{4}{*}{PushObj, seen} & T & 80.7\std{4.2} & 70.2\std{3.7} & 85.1\std{1.4} & 70.9\std{1.7} & 86.0\std{4.0} & 68.9\std{2.7} & 85.3\std{1.3} & 71.8\std{2.8} \\
 & L & 82.7\std{3.1} & 72.2\std{0.4} & 83.8\std{1.7} & 73.1\std{2.0} & 83.3\std{2.3} & 73.3\std{4.2} & 85.3\std{2.4} & 71.6\std{5.3} \\
 & Z & 84.0\std{3.5} & 66.9\std{3.4} & 82.9\std{1.5} & 67.8\std{5.0} & 82.7\std{4.2} & 68.0\std{2.4} & 83.6\std{1.0} & 69.6\std{2.1} \\
 & + & 80.7\std{2.3} & 68.0\std{2.3} & 79.8\std{1.0} & 68.7\std{1.8} & 78.0\std{3.5} & 67.8\std{1.9} & 80.4\std{2.1} & 68.0\std{1.2} \\
\cmidrule(l){2-10}
\multirow{3}{*}{PushObj, unseen} & I & 37.3\std{3.1} & 35.6\std{1.5} & 66.9\std{4.3} & 41.3\std{9.6} & 46.7\std{10.3} & 38.0\std{2.0} & 70.0\std{2.9} & 50.7\std{2.4} \\
 & Small T & 74.7\std{3.1} & 60.9\std{0.4} & 76.9\std{3.4} & 63.6\std{2.1} & 71.3\std{9.0} & 63.6\std{2.8} & 77.8\std{4.5} & 66.0\std{0.7} \\
 & Square & 42.0\std{3.5} & 34.9\std{3.4} & 64.2\std{2.1} & 41.6\std{3.4} & 49.3\std{3.1} & 38.0\std{4.7} & 64.2\std{5.4} & 39.6\std{1.4} \\
\cmidrule(l){2-10}
\multicolumn{2}{@{}l}{PushObj, all shapes} & 68.9\std{0.5} & 58.4\std{1.2} & 77.1\std{0.6} & 61.0\std{2.2} & 71.0\std{2.3} & 59.7\std{0.5} & 78.1\std{0.9} & 62.4\std{1.1} \\
\midrule
\multirow{4}{*}{PointMaze} & Default & 85.3\std{5.0} & 86.0\std{0.7} & 85.6\std{1.5} & 85.6\std{1.0} & 84.0\std{5.3} & 83.3\std{1.3} & 85.3\std{1.2} & 84.0\std{1.2} \\
 & Low mass & 82.0\std{5.3} & 79.8\std{3.2} & 84.0\std{0.7} & 85.8\std{2.7} & 87.3\std{4.2} & 83.8\std{1.7} & 87.6\std{1.5} & 86.0\std{4.0} \\
 & High damping & 82.7\std{3.1} & 84.9\std{1.4} & 85.1\std{0.8} & 83.8\std{1.4} & 82.7\std{2.3} & 84.0\std{0.7} & 83.6\std{1.7} & 85.3\std{3.1} \\
 & Held-out layouts & 67.3\std{3.1} & 74.7\std{1.2} & 77.3\std{2.9} & 82.9\std{2.1} & 57.3\std{6.4} & 69.1\std{6.0} & 74.7\std{2.3} & 77.3\std{2.0} \\
\cmidrule(l){2-10}
\multicolumn{2}{@{}l}{PointMaze, all settings} & 79.3\std{2.1} & 81.3\std{1.1} & 83.0\std{0.6} & 84.5\std{0.9} & 77.8\std{2.9} & 80.1\std{1.8} & 82.8\std{0.7} & 83.2\std{1.0} \\
\bottomrule
\end{tabular*}
\end{table*}

\paragraph{Effects of resetting the fast learner on each task.}
\label{sec:supp_reset}
\ours resets its fast learner at the start of each episode, and \cref{tab:reset} shows that removing the reset lowers success on PushObj.
Here, we further report this comparison for each PushObj shape and for PointMaze (\cref{tab:supp_noreset}).
On PushObj, never resetting lowers success on every shape with both planners, by up to 25.6 points.
On the medium maze, it changes success by at most 3.5 points.
On the held-out layouts, it even raises success, by 7.4 (GD) and 11.8 points (CEM) for Fast only and by 5.6 and 2.6 for \ours.
As mentioned in \cref{sec:ablation}, these layouts recur within a stream, so the carried-over fast weights remain useful.
For \ours, this gain fades over the stream, from 10.0 points in the first 50 episodes to 0.7 in the last 50 (GD), possibly because the slow learner gradually learns these layouts.
How much of the offset is episode-specific may differ across tasks and models, and it is generally unknown before deployment.
Resetting is therefore the safer default.
For example, with \ours, never resetting loses 16.1 (GD) and 15.7 points (CEM) over the seven PushObj shapes, whereas resetting gives up only 5.6 and 2.6 points on the held-out layouts.

\begin{table*}[!htb]
\centering
\caption{\textbf{Effects of limiting the buffer size of the slow learner} (success rate (\%) on PushObj, over all seven shapes).
The first row is \ours as in \cref{tab:main}, which keeps every window.
\emph{Most recent} keeps the newest windows.
\emph{Window reservoir} keeps a uniform random subset of all windows seen so far, by reservoir sampling.
\emph{Episode reservoir} samples whole episodes in the same way.
\emph{Largest error} keeps the windows with the largest prediction error under the current slow learner.
Memory is the size of the buffer after 150 episodes.}
\label{tab:supp_buffer}
\scriptsize
\begin{tabular*}{0.62\textwidth}{@{\extracolsep{\fill}}l cc cc@{}}
\toprule
 & \multicolumn{2}{c}{GD} & \multicolumn{2}{c}{CEM} \\
\cmidrule(lr){2-3} \cmidrule(l){4-5}
Buffer of the slow learner & Success & Memory (MiB) & Success & Memory (MiB) \\
\midrule
All windows (default) & 77.1\std{0.6} & 6.55 & 78.1\std{0.9} & 7.00 \\
\midrule
\multicolumn{5}{@{}l}{At most 2,048 windows} \\
\quad Most recent & 76.2\std{0.5} & 0.79 & 76.9\std{0.7} & 0.78 \\
\quad Window reservoir & 77.3\std{1.0} & 3.41 & 78.3\std{1.0} & 3.50 \\
\quad Episode reservoir & 75.0\std{2.1} & 0.78 & 74.3\std{0.7} & 0.75 \\
\quad Largest error & 75.7\std{0.1} & 1.59 & 77.4\std{0.6} & 1.70 \\
\midrule
\multicolumn{5}{@{}l}{At most 4,096 windows} \\
\quad Most recent & 76.2\std{0.8} & 1.56 & 77.6\std{0.8} & 1.55 \\
\quad Window reservoir & 76.7\std{1.5} & 5.05 & 78.5\std{0.4} & 5.19 \\
\quad Episode reservoir & 76.6\std{0.9} & 1.55 & 77.0\std{0.5} & 1.52 \\
\quad Largest error & 76.3\std{1.1} & 2.85 & 78.0\std{1.1} & 3.07 \\
\bottomrule
\end{tabular*}
\end{table*}

\paragraph{Effects of limiting the buffer size of the slow learner.}
\label{sec:supp_buffer}
By default, \ours keeps the dense windows of every episode in the buffer of its slow learner.
This buffer is small, taking only 6.55 MiB (GD) and 7.00 MiB (CEM) after 150 episodes on PushObj, but it grows with the stream.
Here, we further examine whether its size can be limited for longer deployments.
We keep at most 2,048 or 4,096 windows and evict the rest with one of four policies (\cref{tab:supp_buffer}).
With \emph{Window reservoir}, the success rate stays within 0.4 points of the unlimited buffer at both capacities.
\emph{Most recent} takes much less memory, only 0.78--0.79 MiB with 2,048 windows, and loses 0.9 points (GD) and 1.2 points (CEM).
\emph{Episode reservoir} with 2,048 windows loses the most, 2.1 points (GD) and 3.8 points (CEM), possibly because on average only 58--64 of the 150 episodes ever enter its buffer.
With 4,096 windows, every policy stays within 1.1 points of the unlimited buffer.
The buffer size can thus be limited with little loss.

In \cref{tab:supp_buffer}, buffers with the same number of windows take different amounts of memory.
For example, with 2,048 windows and GD, \emph{Most recent} takes 0.79 MiB and \emph{Window reservoir} 3.41 MiB.
The buffer stores each frame only once, and a window merely points to its frames (\cref{sec:impl_sft}).
Consecutive windows of an episode overlap, so each frame serves several windows.
The unlimited buffer, \emph{Most recent} and \emph{Episode reservoir} keep consecutive windows and store only about 0.4 frames per window.
\emph{Window reservoir} keeps scattered windows, which overlap less, and stores 1.5--2.1 frames per window.
\emph{Largest error} lies in between, with \mbox{0.8--1.0}.

\begin{table*}[!htb]
\centering
\caption{\textbf{Effects of hyperparameters in the slow learner} (success rate (\%) on PushObj with CEM).
Each column changes one setting of the slow learner and keeps the others at their defaults (\cref{tab:impl_sft}).
By default, the LoRA rank is 8 and the slow learner takes two steps per episode.
When the rank changes, $\alpha$ changes with it so that $\alpha/$rank stays at 2.
In the last column, the LoRA is updated before the fast learner is reset instead of after it.
Its loss is then computed with the fast weights of the finished episode in place, and still only the LoRA is updated.
Protocol as in \cref{tab:main}.}
\label{tab:supp_slow}
\scriptsize
\setlength{\tabcolsep}{4pt}
\begin{tabular*}{0.68\textwidth}{@{\extracolsep{\fill}}l c cc cc c@{}}
\toprule
 & & \multicolumn{2}{c}{LoRA rank} & \multicolumn{2}{c}{Slow learner steps} & LoRA update \\
\cmidrule(lr){3-4} \cmidrule(lr){5-6}
Shape & Default & 4 & 16 & 1 & 4 & before fast reset \\
\midrule
T (seen) & 85.3\std{1.3} & 87.8\std{1.5} & 86.9\std{2.7} & 86.0\std{1.8} & 86.0\std{3.1} & 86.0\std{1.8} \\
I (unseen) & 70.0\std{2.9} & 69.6\std{0.8} & 69.3\std{1.8} & 65.8\std{2.7} & 71.8\std{3.8} & 68.0\std{0.7} \\
\bottomrule
\end{tabular*}
\end{table*}

\paragraph{Effects of hyperparameters in the slow learner.}
Here, we further examine how sensitive \ours is to the hyperparameters and one design choice of its slow learner.
We change one setting at a time and rerun the streams of a seen shape (T) and an unseen shape (I) with CEM (\cref{tab:supp_slow}).
The LoRA rank has little effect.
Ranks 4 and 16 instead of 8 change the success rate by at most 2.5 points on T and 0.7 points on I.
The number of steps the slow learner takes per episode matters mainly on the unseen shape.
On I, one step instead of two loses 4.2 points and four steps gain 1.8 points, whereas T changes by 0.7 points in both cases.
By default, we reset the fast learner before the slow learner updates, so the slow learner's loss is computed with $\tz$+LoRA.
Here, we also test the opposite order, where this loss is computed with the fast weights of the finished episode still in place.
The fast weights are kept fixed in these steps, and only the LoRA is updated.
This changes the success rate by $+0.7$ points on T and $-2.0$ points on I.
Overall, \ours does not rely on a particular setting of its slow learner.
Every variant in \cref{tab:supp_slow} still outperforms all baselines in \cref{tab:supp_pertask}, by at least 6.7 points on T and 12.9 points on I.

\end{document}